\documentclass[10pt,journal]{IEEEtran}

\def\BibTeX{{\rm B\kern-.05em{\sc i\kern-.025em b}\kern-.08em
    T\kern-.1667em\lower.7ex\hbox{E}\kern-.125emX}}

\usepackage{dsfont}
\usepackage{bm}

\usepackage{times}
\usepackage{amssymb}
\usepackage{amsmath}
\usepackage{algpseudocode}
\usepackage{algorithm}
\usepackage{graphicx}
\usepackage{subfigure}
\usepackage{array}
\usepackage{multirow}
\usepackage{makecell}
\usepackage{subfigure}
\usepackage{caption}

\usepackage{tikz}
\usetikzlibrary{arrows}

\algnewcommand\algorithmicforeach{\textbf{for each}}
\algdef{S}[FOR]{ForEach}[1]{\algorithmicforeach\ #1\ \algorithmicdo}

\newcommand{\be}{\begin{equation}}
\newcommand{\ee}{\end{equation}}
\newcommand{\ba}{\begin{eqnarray}}
\newcommand{\ea}{\end{eqnarray}}

\newcommand{\norm}[1]{\left\lVert#1\right\rVert}
\newcommand{\qedb}{\hfill\ensuremath{\blacksquare}}

\input{epsf}

\begin{document}

\title{Twin Sorting Dynamic Programming Assisted User Association and Wireless Bandwidth Allocation for Hierarchical Federated Learning} 

\author{Rung-Hung~Gau,
	Ting-Yu Wang
        and~Chun-Hung~Liu
\thanks{This article was presented in part at the IEEE VTC-Spring 2024.}
\IEEEcompsocitemizethanks{
	\IEEEcompsocthanksitem Rung-Hung Gau is with Institute of Communications Engineering, National Yang Ming Chiao Tung University, Hsinchu, Taiwan\\ 
	(e-mail: gaurunghung@nycu.edu.tw).
	\IEEEcompsocthanksitem Ting-Yu Wang is with Institute of Communications Engineering, National Yang Ming Chiao Tung University, Hsinchu, Taiwan\\ 
	(e-mail: wty880113.ee06@nycu.edu.tw).
	\IEEEcompsocthanksitem Chun-Hung Liu is with Department of Electrical and Computer Engineering, Mississippi State University, Mississippi State, MS, USA\\ 
	(e-mail: chliu@utexas.edu).}
}

\maketitle

\begin{abstract}
In this paper, we study user association and wireless bandwidth allocation for a hierarchical federated learning system that consists of mobile users, edge servers, and a cloud server. To minimize the length of a global round in hierarchical federated learning with equal bandwidth allocation, we formulate a combinatorial optimization problem. We design the twin sorting dynamic programming (TSDP) algorithm that obtains a globally optimal solution in polynomial time when there are two edge servers. In addition, we put forward the TSDP-assisted algorithm for user association when there are three or more edge servers. Furthermore, given a user association matrix, we formulate and solve a convex optimization problem for optimal wireless bandwidth allocation. Simulation results show that the proposed approach outperforms a number of alternative schemes.  
\end{abstract}


\section{Introduction}

Federated learning \cite{McMahan2017} is a decentralized machine learning framework designed to benefit from parallel computation of user devices and maintain data privacy. In a basic federated learning system, each device directly uploads its local model of machine learning to the cloud/parameter server. Hierarchical federated learning (HFL)  \cite{Abad2020ICASSP} \cite{Liu2020ICC} was proposed to improve the scalability of federated learning. Abad \textit{et al.} \cite{Abad2020ICASSP} investigated HFL in a cellular network composed of a macro base station, small base stations, and mobile devices. In this paper, we study a client-edge-cloud HFL system comprising mobile devices, edge servers, and a cloud server \cite{Liu2020ICC}. Specifically, a mobile device is associated with an edge server. An edge server takes charge of forwarding the latest global model to associated mobile devices. In addition, an edge server is responsible for aggregating models of corresponding mobile devices and sending the results to the cloud server. 

A typical federated learning process consists of several rounds. For a client-edge-cloud HFL process, a round is composed of three phases. In the first phase, each mobile device adopts the received global model and its data to perform local model updates and transmits the updated local model to the associated edge server. In the second phase, an edge server aggregates the received local models, obtains the edge model, and sends the edge model to the cloud server. In the third phase, the cloud server updates the global model based on the received edge models and broadcasts the latest global model to edge servers which in turn forward the global model to corresponding mobile devices. 

User association and wireless bandwidth allocation are essential for optimizing the performance of HFL, especially when mobile clients have unequal computation capabilities and edge servers have different model uploading delays to the cloud server. Liu \textit{et al.} \cite{Liu2020ICC} proposed a scheme that assigns the same number of clients to all edge servers in a client-edge-cloud HFL system. In contrast, we aim to optimally assign clients to edge servers to minimize the HFL latency, defined as the length of a global round of HFL. Luo \textit{et al.} \cite{Luo2020TWC} proposed using device transferring adjustments and device exchanging adjustments to select an adequate user association matrix for HFL. Denote the latency for mobile user $m$ to update and upload its local model to the associated edge server by $u_{m}$, $\forall m$. Denote the latency for edge server $n$ to upload its model to the cloud server by $v_{n}$, $\forall n$. Liu \textit{et al.} \cite{Liu2022TWC} aimed at jointly optimizing user association and wireless resource allocation for minimizing a linear combination of $\max_{m} u_{m}$ and $\max_{n} v_{n}$ in wireless HFL. They proposed the Max-SNR algorithm that assigns a mobile user to the edge server with the largest signal-to-noise ratio (SNR). In this paper, we focus on minimizing the HFL latency, which is not a linear combination of $\max_{m} u_{m}$ and $\max_{n} v_{n}$. Liu \textit{et al.} \cite{Liu2022SEC} sought to optimize the user-edge association matrix and the  transmission power vector of mobile users in an HFL system. They proposed a greedy algorithm that assigns users to edge servers based on signal-to-noise ratios. In \cite{Liu2022SEC}, the bandwidth allocated to a mobile user is fixed and given in advance. In this paper, we study the case in which mobile users compete for the bandwidth of wireless communications. Specifically, mobile users associated with the same edge server equally share the bandwidth allocated to the edge server. Namely, equal bandwidth allocation (EBA) is adopted. For the scenario studied in the paper, the Max-SNR algorithm and the above greedy algorithm do not produce optimal solutions.

In a large-scale HFL system, edge-to-cloud delays could be as large as or greater than mobile-to-edge delays, especially when edge servers are geographically far away from the cloud server or the path from an edge server to the cloud server contains a congested Internet router. Gau \textit{et al.} \cite{Gau2024VTC} designed the backbone-aware greedy (BAG) algorithm for finding an adequate matrix of user association based on the computational capabilities of mobile users, wireless communication delays, and edge-to-cloud delays. Specifically, they aimed to minimize the HFL latency. While the BAG algorithm does not always produce an optimal solution, we put forward the twin sorting dynamic programming (TSDP) algorithm that always obtains an \textit{optimal} matrix of user association in \textit{polynomial time} when an HFL system contains two edge servers and adopts equal bandwidth allocation for connecting mobile users to edge servers.

Reference \cite{Wu2023TPDS} put forward using deep reinforcement learning based staleness control and heterogeneity-aware client-edge association for improving the system efficiency of HFL. To accelerate the HFL process, Wang \textit{et al.} \cite{Wang2023TMC} proposed FedCH that selects cluster heads and utilizes a bipartite matching algorithm to find the cluster head for each mobile device based on the computational capability and the communication latency. The heterogeneity-aware client-edge association algorithm \cite{Wu2023TPDS} and the cluster construction algorithm \cite{Wang2023TMC} assumed that the amount of bandwidth allocated to a mobile device is fixed and the mobile-edge communication latency does not change as the number of mobile devices associated with an edge server increases. In contrast, we study the case in which the communication latency between a mobile device and an edge server depends on the number of mobile devices that connect to the edge server. Deng \textit{et al.} \cite{Deng2024ToN} sought to minimize the total communication cost for HFL by optimally selecting edge aggregators and mobile-edge associations. As \cite{Wu2023TPDS} \cite{Wang2023TMC}, they assumed that the communication cost between a mobile device and an edge aggregator is given and does not depend on the number of mobile devices connecting to the edge aggregator.

The convergence of a synchronized federated learning process has been extensively studied in the literature \cite{Dinh2021ToN} \cite{Chen2021TWC}. Dinh \textit{et al.} \cite{Dinh2021ToN} proposed a wireless federating learning algorithm that is able to handle heterogeneous user equipment (UE) data. In addition, they obtained the convergence rate that characterizes the trade-off between the local computation rounds of each UE and the global computation rounds. Under some assumptions, Chen \textit{et al.} \cite{Chen2021TWC} proved that $\mathbb{E}[F(\mathbf{g}_{k+1})-F(\mathbf{g}^{*})] \le a_{k}+b_{k} \mathbb{E}[F(\mathbf{g}_{k})-F(\mathbf{g}^{*})]$, where $\mathbb{E}$ represents expectation, $F$ is the loss function, $\mathbf{g}_{k}$ is the global model in round $k$, $\mathbf{g}^{*}$ is the optimal global model and $(a_{k},b_{k})$ are two real numbers, $\forall k$. The required time for an HFL process to converge typically depends on the user association policy. However, since the global model is only updated in the end of each round, changing the value of the user association matrix does not alter the required number of rounds for a synchronized federated learning process to converge. When the number of rounds required for convergence is fixed, minimizing the time for convergence is equivalent to minimizing the average length of a round. We seek to minimize the length of each round through optimal user association in this paper.

Wen \textit{et al.} \cite{Wen2022TWC} obtained solutions for the problem of bit and sub-channel allocation and the problem of helper scheduling in a hierarchical federated learning system. Liu \textit{et al.} \cite{Liu2023TWC} studied HFL with neural network quantization. They derived a tighter convergence bound and optimized the two aggregation intervals for HFL. Nevertheless, they \cite{Wen2022TWC} \cite{Liu2023TWC} did not tackle the user association problem. Chen \textit{et al.} \cite{Chen2023TVT} put forward a deep reinforcement learning approach for adapting the device selection and resource allocation strategies in asynchronous HFL systems. Feng \textit{et al.} \cite{Feng2022TWC} proposed a mobility-aware cluster federated learning algorithm. Client selection and mobility management for HFL are beyond the scope of this paper.

Our major technical contributions include the following.

\begin{itemize}

    \item For hierarchical federated learning with equal bandwidth allocation, we formulate a combinatorial optimization problem to obtain an optimal user association matrix that minimizes the length of a global round. 
    
    \item We put forward the twin sorting dynamic programming algorithm that produces an optimal user association matrix in polynomial time when there are two edge servers and equal bandwidth allocation is adopted in an HFL system. To the best of our knowledge, the proposed TSDP algorithm is the \textit{first polynomial-time} algorithm that solves the above optimal user association problem with two edge servers. 
    
    \item For an HFL system that contains three or more edge servers, we propose using the TSDP-assisted algorithm to obtain an adequate user association matrix in polynomial time. 
    
    \item Given a user association matrix, we formulate and solve a convex optimization problem for achieving optimal wireless bandwidth allocation that further reduces the HFL latency. 
    
    \item We use simulations to demonstrate that the proposed approach could significantly outperform a number of alternative schemes in the literature. 
    
\end{itemize}

The rest of the paper is organized as follows. We briefly introduce related work on user association in Section II. In Section III, we include the system models and formulate a combinatorial optimization problem for backbone-aware user association in hierarchical federated learning with equal bandwidth allocation. In Section IV, we elaborate on the proposed twin sorting dynamic programming algorithm that finds an optimal matrix of user association in polynomial time when there are two edge servers in the studied HFL system. In Section V, we put forward the TSDP-assisted algorithm for obtaining an adequate matrix of user association when there are more than two edge servers. In addition, the TSDP-assisted algorithm formulates and solves convex optimization problems for optimal wireless bandwidth allocation. In Section VI, we include simulation results that reveal the advantages of the proposed approaches. Our conclusions for this study are included in Section VII. 

\section{Related Work}

Even before the invention of federated learning, user association plays an essential role in a wireless communication network consisting of multiple base stations. To optimize the system performance, it is important for users to connect to optimal base stations. Fooladivanda \textit{et al.} \cite{Fooladivanda2013TWC} investigated joint user association and resource allocation for heterogeneous networks (HetNets). Ye \textit{et al.} \cite{Ye2013TWC} studied user association for load balancing in a HetNet and sought to maximize the aggregated utility function. Lin \textit{et al.} \cite{Lin2015JSAC} adopted stochastic geometry and optimization theory to analytically obtain the optimal user association bias factors and spectrum partition ratios for multi-tier HetNets. To deal with interference mitigation, user association, and resource allocation in HetNets, Oo \textit{et al.} \cite{Oo2017TMC} proposed using two approaches: Markov approximation and payoff-based log-linear learning. The first approach is based on a novel Markov chain design, while the second approach is based on game theory. Zhao \textit{et al.} \cite{Zhao2019TWC} put forward using multi-agent deep reinforcement learning for a distributed optimization of user association and resource allocation in HetNets. 

User association has been jointly optimized with several advanced technologies for wireless communications, such as cloud radio access network (C-RAN) \cite{Luo2015TWC}, Multiple-Input Multiple-Output (MIMO)  \cite{Chien2016TWC}, and full-duplex that allows concurrent communication in both directions \cite{Sekander2017TMC}. While many earlier works on wireless communications focused on microwaves, recent works exploited millimeter waves to increase the data transmission rate. Khawam \textit{et al.} \cite{Khawam2022TMC} utilized non-cooperative game theory for coordinated user association and spectrum allocation in 5G HetNet with microwave and millimeter wave. Zarifneshat \textit{et al.} \cite{Zarifneshat2023TMC} took a bi-objective optimization approach for user association in millimeter wave cellular networks. Specifically, the two objectives depend on the base station utility and the blockage score. Huang \textit{et al.} \cite{Huang2022TMC} formulated the problem of online user association and resource allocation in a wireless caching network as a stochastic network optimization problem. In addition, they designed the Predictive User-AP Association and Resource Allocation (PUARA) scheme that achieves a provably near-optimal throughput with queue stability. Li \textit{et al.} \cite{Li2022TMC} studied the problem of jointly optimizing content caching and user association for edge computing in HetNets and proved that the problem is NP-hard. Chen \textit{et al.} \cite{Chen2023TMC} investigated cache placement, video quality decision, and user association for live video streaming in mobile edge computing systems. Nouri \textit{et al.} \cite{Nouri2023TMC} proposed an efficient algorithm based on machine learning to optimize the uncrewed aerial vehicle (UAV) placements and the user association in wireless MIMO networks where UAVs serve as aerial base stations. Dai \textit{et al.} \cite{Dai2023TMC} put forward downlink-uplink decoupling with which each user equipment (UE) is allowed to associate with different UAVs for downlink and uplink transmissions in UAV networks. To maximize the sum rate, they adopted a partially observable Markov decision process (POMDP) model and proposed a multi-agent deep reinforcement learning approach that enables each UAV to choose its policy in a distributed manner. These schemes were not designed for hierarchical federated learning. 

Liu \textit{et al.} \cite{Liu2020TVT} aimed to maximize a mobile communication network's long-term average communication efficiency by selecting an optimal pair of base station and network slice for each user/device in each time slot. They proposed an efficient device association scheme for radio access network slicing based on deep reinforcement learning and federated learning. Lim \textit{et al.} \cite{Lim2021JSAC} proposed a hierarchical game theoretic framework for edge association and resource allocation in self-organizing HFL networks. Li \textit{et al.} \cite{Li2022TVT} put forward a deep reinforcement learning based approach for the cloud server to decide edge association in cluster-based personalized federated learning systems. Lin \textit{et al.} \cite{Lin2023TVT} adopted federated multi-agent reinforcement learning to tackle privacy-preserving edge association and power allocation for the Internet of Vehicles. Instead of using machine learning or game theory, we adopt dynamic programming and sorting to minimize the HFL latency. Specifically, we prove that our proposed algorithm always finds a globally optimal solution in polynomial time for the studied user association problem when there are two edge servers in the HFL system. 

Ong \textit{et al.} \cite{Ong2022ICC} put forward using local losses to dynamically select clients for federated learning systems. Hsu \textit{et al.} \cite{Hsu2023VTC} designed an MMSE-based power control scheme for wireless federated learning. Nevertheless, they \cite{Ong2022ICC} \cite{Hsu2023VTC} did not study user-edge association for HFL. Hosseinalipour \textit{et al.} \cite{Hosseinalipour2022ToN} studied hierarchical federated learning in which clusters could be formed via device-to-device (D2D) communications and developed a distributed algorithm to tune the D2D rounds in each cluster. They assumed that the hierarchical structure was given and did not address the user association problem. Ganguly \textit{et al.} \cite{Ganguly2023ToN} considered FL in a three-tier wireless network and proposed using a dynamically selected edge server to aggregate models of machine learning. Liu \textit{et al.} \cite{Liu2024ToN} put forward a layer-wise aggregation mechanism for decentralized federated learning that does not have a centralized parameter server and utilizes peer-to-peer (P2P) communications. Decentralized federated learning based on P2P communications is beyond the scope of this paper. Reference \cite{Lim2020Survey} comprehensively surveyed federated learning in mobile edge networks. Reference \cite{Kar2023Survey} contained a survey of offloading in federated cloud–edge–fog systems.

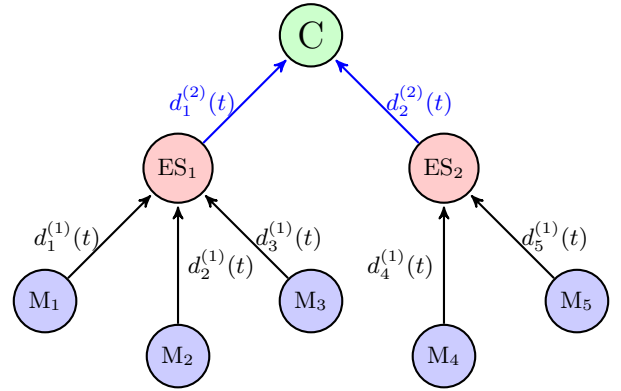
\begin{figure}[b]
\begin{center}

\begin{tikzpicture}[->,>=stealth',shorten >=1pt,auto,node distance=2.5cm,
  thick, cloud node/.style={circle,fill=green!20,draw,font=\sffamily\Large\bfseries},
  thick, edge node/.style={circle,fill=red!20,draw,font=\sffamily\small\bfseries},
  thick, mobile node/.style={circle,fill=blue!20,draw,font=\sffamily\small\bfseries}]
  
  \node[cloud node] (1) {$\mathrm{C}$};
  \node[edge node] (2) [below left of=1] {$\mathrm{ES}_{1}$};
  \node[edge node] (3) [below right of=1] {$\mathrm{ES}_{2}$};
  \node[mobile node] (4) [below left of=2] {$\mathrm{M}_{1}$};
  \node[mobile node] (5) [below of=2] {$\mathrm{M}_{2}$};
  \node[mobile node] (6) [below left of=3] {$\mathrm{M}_{3}$};
  \node[mobile node] (7) [below of=3] {$\mathrm{M}_{4}$};
  \node[mobile node] (8) [below right of=3] {$\mathrm{M}_{5}$};
         
  \path[every node/.style={font=\sffamily\small}]
    (2) edge [blue] node [left] {$d_{1}^{(2)}(t)$} (1)
    (3) edge [blue] node [right] {$d_{2}^{(2)}(t)$} (1)
    (4) edge [black] node [left] {$d_{1}^{(1)}(t)$} (2)
    (5)  edge [black] node [right] {$d_{2}^{(1)}(t)$} (2)
    (6) edge [black] node [right] {$d_{3}^{(1)}(t)$} (2)
    (7)	 edge [black] node [left] {$d_{4}^{(1)}(t)$} (3)
    (8) edge [black] node [right] {$d_{5}^{(1)}(t)$} (3);

\end{tikzpicture}    

	\caption{A hierarchical federated learning system in round $t$.}
	\label{SystemModel}
\end{center}	
\end{figure}

\section{System Models and Problem Formulation}

A hierarchical federated learning system contains one cloud server, $N \ge 2$ edge servers and $M \ge 2$ mobile devices. Edge server $n$ is co-located with base station (BS) $n$, $\forall n$. Let $\mathbb{N}=\{1,2,3,...\}$ be the set of natural numbers. For each $n \in \mathbb{N}$, let $[n]=\{1,2,..,n\}$. Denote the cardinality of a set $S$ by $|S|$. We reuse the notations in \cite{Gau2024VTC} whenever appropriate.

A hierarchical federated learning process consists of multiple rounds. A mobile device is associated with an edge server in each round. Let $x_{m,n}(t) \in \{0,1\}$ be a binary variable, $\forall m \in [M], n \in [N], t \in \mathbb{N}$. If mobile device $m$ is associated with edge server $n$ in round $t$, $x_{m,n}(t)=1$. Otherwise, $x_{m,n}(t)=0$. Each mobile device keeps a machine learning model. If $x_{m,n}(t)=1$, mobile device $m$ sends its model of machine learning to edge server $n$ via BS $n$ in round $t$. For the reason that a mobile device is associated with only one edge server in an HFL round, we have 
\ba
\sum_{n=1}^{N} x_{m,n}(t)=1, \forall m \in [M],t \in \mathbb{N}.
\ea

Let $A_{n}(t)$ be the set consisting of the indexes of mobile devices associated with edge server $n$ in round $t$ of HFL, $\forall n \in [N], t \in \mathbb{N}$. Specifically,
\ba
A_{n}(t)=\{m \in [M] | x_{m,n}(t)=1\}, \forall n \in [N], t \in \mathbb{N}.
\ea
Note that $\cup_{n=1}^{N} A_{n}(t)=[M]$ and $A_{i}(t) \cap A_{j}(t)=\emptyset$, $\forall i \ne j$. Namely, $(A_{1}(t),A_{2}(t),..,A_{N}(t))$ is a partition of $[M]$. Let $A(t)=(A_{1}(t),A_{2}(t),..,A_{N}(t))$.

The cloud server maintains the global model that typically changes with time until convergence. Denote the global model vector at the beginning of round $t$ by $\mathbf{w}(t)$, $\forall t \in \mathbb{N}$. At the beginning of round $t$, the cloud server transmits the value of $\mathbf{w}(t)$ to all mobile devices via the base stations in the HFL system. Once a mobile device acquires the value of $\mathbf{w}(t)$, it utilizes $\mathbf{w}(t)$ and local training data to update the local model. Define $\alpha_{m}(t)$ as the required time for mobile device $m$ to accomplish a local model update in round $t$, $\forall m \in [M], t \in \mathbb{N}$. When mobile device $m$ is equipped with a graphics processing unit (GPU), $\alpha_{m}(t)$ depends on the computing capability of the GPU and the amount of training data.

The mobile devices that connect to the same edge server use orthogonal channels. Let $B_{n}$ be the amount of wireless bandwidth of which BS/edge server $n$ takes charge. Let $\theta_{m,n}(t)$ be the fraction of bandwidth that edge server $n$ assigns to mobile device $m$ in HFL round $t$, $\forall m \in [M], n \in [N], t \in \mathbb{N}$. If $x_{m,n}(t)=0$, mobile device $m$ does not connect to edge server $n$ and therefore $\theta_{m,n}(t)=0$. Otherwise, $\theta_{m,n}(t)>0$. In addition, $\sum_{m=1}^{M} \theta_{m,n}(t) \le 1$, $\forall n \in [N], t \in \mathbb{N}$. Denote the transmit power of mobile device $m$ in HFL round $t$ by $p_{m}(t)$. Let $g_{m,n}(t)$ be the channel gain of the wireless link from mobile device $m$ to BS/edge server $n$ in HFL round $t$. Let $N_{0}$ be the power spectral density of the additive white Gaussian noise at each edge server. Let $r_{m,n}(t)$ be the data transmission rate from mobile device $m$ to BS/edge server $n$ in HFL round $t$. Based on information theory, 
\ba
r_{m,n}(t) &=& \theta_{m,n}(t) B_{n} \times \log_{2}(1+\frac{p_{m}(t) g_{m,n}(t)}{\theta_{m,n}(t) B_{n} N_{0}}), \nonumber\\
&& \forall m \in [M], n \in [N], t \in \mathbb{N}.
\ea

Let $L$ be the number of bits required for representing a local model of machine learning. Let $\beta_{m,n}(t)$ be the amount of time it takes to upload the local model of mobile device $m$ to edge server $n$ in round $t$ when $\theta_{m,n}(t)=1$, $\forall m \in [M], n \in [N], t \in \mathbb{N}$. In other words,
\ba
\beta_{m,n}(t) &=& \frac{L}{B_{n} \times \log_{2}(1+\frac{p_{m}(t) g_{m,n}(t)}{B_{n} N_{0}})}.
\ea

In each HFL round, mobile devices that are associated with the same edge server equally share the corresponding wireless bandwidth. Namely, equal bandwidth allocation (EBA) is used and
\ba
\theta_{m,n}(t) &=& \frac{x_{m,n}(t)}{|A_{n}(t)|}, \forall m \in [M], n \in [N], t \in \mathbb{N}.
\ea

Let $t_{m,n}(t)$ be the amount of time that mobile device $m$ takes to transmit the local model to edge server $n$ if the former is associated with the latter in round $t$. Specifically,
\ba
t_{m,n}(t) &=& \frac{L}{r_{m,n}(t)} \nonumber\\
&=& \frac{L}{\theta_{m,n}(t) B_{n} \times \log_{2}(1+\frac{p_{m}(t) g_{m,n}(t)}{\theta_{m,n}(t) B_{n} N_{0}})} \nonumber\\
&=& \frac{|A_{n}(t)| \cdot L}{B_{n} \times \log_{2}(1+ \frac{p_{m}(t) g_{m,n}(t)}{B_{n} N_{0}/|A_{n}(t)|})} \nonumber\\
&\le& |A_{n}(t)| \cdot \beta_{m,n}(t).
\ea

Let $d_{m}^{(1)}(t)$ be the amount of time required for mobile device $m$ to carry out local model updates and send the latest local model to the associated edge server in round $t$, $\forall m \in [M], t \in \mathbb{N}$. Then, 
\ba
d_{m}^{(1)}(t) &=& \alpha_{m}(t)+\sum_{k=1}^{N} x_{m,k}(t) \times \beta_{m,k}(t) \times |A_{k}(t)|, \nonumber\\
&& \forall m \in [M], t \in \mathbb{N}.
\label{DelayDevice2Edge}
\ea 

Let $d_{n}^{(2)}(t)$ be the compulsory amount of time for edge server $n$ to transmit its model to the cloud server, $\forall n \in [N], t \in \mathbb{N}$. In general, the value of $d_{n}^{(2)}(t)$ depends on $n$ and $t$ mainly due to the following reasons. First, the geographical distance between edge server $n$ and the cloud server typically changes with $n$. In addition, the end-to-end delay between edge server $n$ and the cloud server depends on the time-varying queueing delay of intermediate Internet routers.  

In Fig. \ref{SystemModel}, we illustrate an HFL system that is composed of five mobile devices, two edge servers and a cloud server. Specifically, the cloud server is represented by a green circle marked by $\mathrm{C}$, the $n$th edge server is depicted by a red circle marked by $\mathrm{ES}_{n}$, $\forall n \in [N]$ and the $m$th mobile device is portrayed by a blue circle marked by $\mathrm{M}_{m}$, $\forall m \in [M]$. Moreover, the first three mobile devices connect to the first edge server, while the remaining two mobile devices are associated with the second edge server in round $t$. In this case, $x_{1,1}(t)=x_{2,1}(t)=x_{3,1}(t)=1$ and $x_{4,2}(t)=x_{5,2}(t)=1$. Meanwhile, $A_{1}(t)=\{1,2,3\}$ and $A_{2}(t)=\{4,5\}$.

Denote the length of HFL round $t$ by $y(t)$, $\forall t \in \mathbb{N}$. Then,
\ba
y(t) = \max_{n: n \in [N]} \big[\max_{m: m \in A_{n}(t)} d_{m}^{(1)}(t)+d_{n}^{(2)}(t)\big], \forall t \in \mathbb{N}.
\ea

We now elaborate on the above equation. Consider HFL round $t$. Since edge server $n$ has to collect models from the associated mobile devices with indexes in $A_{n}(t)$, it takes $\max_{m: m \in A_{n}(t)} d_{m}^{(1)}(t)$ time units  to obtain the latest model of edge server $n$, $\forall n \in [N]$. In addition, edge server $n$ has to spend $d_{n}^{(2)}(t)$ time units to transmit its model to the cloud server. Thus, the cloud server has to wait for $\max_{m: m \in A_{n}(t)} d_{m}^{(1)}(t)+d_{n}^{(2)}(t)$ time units to acquire the latest model of edge server $n$. Since the cloud server has to collect all edge models in order to update the global model, $y(t)$ is equal to $\max_{n: n \in [N]} \big[\max_{m: m \in A_{n}(t)} d_{m}^{(1)}(t)+d_{n}^{(2)}(t)\big]$. 

Let $\mathbf{X}(t) \in \{0,1\}^{M \times N}$ be the user association matrix in HFL round $t$, $\forall t \in \mathbb{N}$. Specifically, $[\mathbf{X}(t)]_{m,n}=x_{m,n}(t)$, $\forall m \in [M], n \in [N]$. To obtain an optimal user association matrix in HFL round $t$, we formulate a combinatorial optimization problem as follows.
\ba
&& \min_{\mathbf{X}(t) \in \{0,1\}^{M \times N}} \max_{n: n \in [N]} \big[\max_{m: m \in A_{n}(t)} d_{m}^{(1)}(t)+d_{n}^{(2)}(t)\big] \nonumber\\
&& \textrm{subject to} \nonumber\\
&& [\mathbf{X}(t)]_{m,n} \in \{0,1\}, \forall m \in [M], n \in [N] \nonumber\\
&& \sum_{n=1}^{N} [\mathbf{X}(t)]_{m,n}=1, \forall m \in [M] \nonumber\\
&& A_{n}(t)=\{m \in [M] | [\mathbf{X}(t)]_{m,n}=1\}, \forall n \in [N].
\label{OptimalUserAssociationVersion1}
\ea
Let $h^{*}$ be the minimum value of the objective function and $\mathbf{X}^{*}(t)$ be an optimal solution of (\ref{OptimalUserAssociationVersion1}). Denote the set of all feasible solutions of (\ref{OptimalUserAssociationVersion1}) by $\Omega_{X}$. Since each of the $M$ mobile devices could be assigned to one of the $N$ edge servers, $|\Omega_{X}|=N^{M}$.

Let $\alpha(t)=(\alpha_{1}(t),\alpha_{2}(t),..,\alpha_{M}(t))$, $\forall t \in \mathbb{N}$. For each $t \in \mathbb{N}$, let $\beta(t) \in \mathbb{R}^{M \times N}$ be a matrix such that the element in the $m$th row, and the $n$th column is $\beta_{m,n}(t)$, $\forall m \in [M],n \in [N]$. Define $d^{(1)}(t)=(d_{1}^{(1)}(t),d_{2}^{(1)}(t),..,d_{M}^{(1)}(t))$ and $d^{(2)}(t)=(d_{1}^{(2)}(t),d_{2}^{(2)}(t),..,d_{N}^{(2)}(t))$. When the HFL round index $t$ is clear from the context, we abbreviate $\alpha_{m}(t)$, $\beta_{m,n}(t)$, $d_{m}^{(1)}(t)$, $d_{n}^{(2)}(t)$ and $A_{n}(t)$ by $\alpha_{m}$, $\beta_{m,n}$, $d_{m}^{(1)}$, $d_{n}^{(2)}$ and $A_{n}$, respectively. In addition, $\alpha(t)$, $\beta(t)$, $d^{(1)}(t)$ and $d^{(2)}(t)$ are abbreviated by $\alpha$, $\beta$, $d^{(1)}$ and $d^{(2)}$, respectively. Furthermore, $\mathbf{X}(t)$ is abbreviated by $\mathbf{X}$, while $\mathbf{X}^{*}(t)$ is abbreviated by $\mathbf{X}^{*}$. 

\begin{algorithm}
\captionsetup{labelsep=period}
\caption{\texttt{TSDP}: twin sorting dynamic programming}
\label{Algorithm-TSDP}
    \begin{algorithmic}[1]
    \Require $M$, $\alpha$, $\beta$, $d^{(2)}$. 
    \Ensure $h^{*}$, $(A_{1}^{*},A_{2}^{*})$, $\mathbf{X}^{*}$.
    \State $h^{*} \gets \max_{m: m \in [M]} \alpha_{m}+M \cdot \beta_{m,2}+d_{2}^{(2)}$. // $h(0)$
    \State $A_{1} \gets \emptyset$, $A_{2} \gets [M]$.
    \For{$k=1$ to $M$}
    	\State $\phi_{k,m}^{(1)} \gets \alpha_{m}+k \cdot \beta_{m,1}$, $\forall m \in [M]$.
	\State $\gamma_{k}^{(1)} \gets \texttt{SortDecreasing}(\phi_{k,1}^{(1)},\phi_{k,2}^{(1)},..,\phi_{k,M}^{(1)})$.
    	\For{$r=1$ to $M$}
		\State Find $s_{1}(k,r)$ based on $\gamma_{k}^{(1)}$ and (\ref{Group1Leader}).
		\State $\Lambda_{k,r} \gets \{m \in [M]| \gamma_{k,m}^{(1)}>r\}$.
		\State $\phi_{M-k,m}^{(2)} \gets \alpha_{m}+(M-k) \cdot \beta_{m,2}$, $\forall m \in \Lambda_{k,r}$.
		\State $u_{k,r} \gets (\phi_{M-k,\Lambda_{k,r}[1]}^{(2)},..,\phi_{M-k,\Lambda_{k,r}[M-r]}^{(2)})$.
         	\State $\gamma_{k,r}^{(2)} \gets \texttt{SortIncreasing}(u_{k,r},\Lambda_{k,r})$.
		\State Obtain $(\Xi_{k,r}^{(1)},\Xi_{k,r}^{(2)})$ based on (\ref{XiSet}).
		\State Obtain $(\zeta_{k,r}^{(1)},\zeta_{k,r}^{(2)})$ based on (\ref{zeta1}) and (\ref{zeta2}).
		\State $h(k,r) \gets \max_{n: n \in [2]} \zeta_{k,r}^{(n)}+d_{n}^{(2)}$.
		\If{$h(k,r)<h^{*}$}
			\State $h^{*} \gets h(k,r)$, $(A_{1}^{*},A_{2}^{*}) \gets (\Xi_{k,r}^{(1)},\Xi_{k,r}^{(2)})$.
		\EndIf
	\EndFor	
     \EndFor
     \For{$m=1$ to $M$}
     	\If{$m \in A_{1}^{*}$}
		\State $[\mathbf{X}^{*}]_{m,1} \gets 1$, $[\mathbf{X}^{*}]_{m,2} \gets 0$.
	\Else
		\State $[\mathbf{X}^{*}]_{m,1} \gets 0$, $[\mathbf{X}^{*}]_{m,2} \gets 1$.
	\EndIf
     \EndFor
    \end{algorithmic}
\end{algorithm}    

\section{The twin sorting dynamic programming algorithm}
 
We consider the case in which $N=2$ in this section. We will deal with the case in which $N \ge 3$ in the next section. We propose the twin sorting dynamic programming (TSDP) algorithm that obtains an optimal solution for (\ref{OptimalUserAssociationVersion1}) when $N=2$ in polynomial time. The proposed algorithm utilizes two similar sorting procedures and dynamic programming. Specifically, the first sorting arranges the latencies of all mobile users in decreasing order from the viewpoint of edge server 1, while the second sorting lists the latencies of a subset of mobile users in increasing order from the viewpoint of edge server 2. Consider HFL round $t \in \mathbb{N}$. Pseudo codes for the TSDP algorithm are included in Algorithm \ref{Algorithm-TSDP}.

We introduce some key ideas and variables for the TSDP algorithm as follows. First, for each $n \in [2]$, based on (\ref{DelayDevice2Edge}), we have
\ba
\max_{m \in A_{n}} d_{m}^{(1)}+d_{n}^{(2)} = \max_{m \in A_{n}} \alpha_{m}+|A_{n}| \cdot \beta_{m,n}+d_{n}^{(2)}.
\label{EdgeLatency}
\ea
When $N=2$ and $t$ is omitted, (\ref{OptimalUserAssociationVersion1}) is equivalent to the subsequent optimization problem.
\ba
&& \min_{(A_{1},A_{2})} \max_{n \in [2]} \big[\max_{m: m \in A_{n}} d_{m}^{(1)}+d_{n}^{(2)} \big] \nonumber\\
&& \textrm{subject to} \nonumber\\
&& A_{1} \cup A_{2}=[M] \nonumber\\
&& A_{1} \cap A_{2}=\emptyset.
\label{OptimalUserAssociationTwoEdgeServers}
\ea
Let $\Omega_{2}$ be the set that is composed of all feasible solutions of (\ref{OptimalUserAssociationTwoEdgeServers}). 

One key idea behind the proposed TSDP algorithm is to partition $\Omega_{2}$ into a finite number of classes and efficiently solve a subproblem associated each class. The TSDP algorithm utilizes two types of sorting. It uses the first type of sorting to partition $\Omega_{2}$ and adopts the second type of sorting to efficiently solve each subproblem. The TSDP algorithm acquires an optimal solution of (\ref{OptimalUserAssociationTwoEdgeServers}) based on the optimal solutions of the subproblems.

To sort mobile users from the viewpoint of edge server $n$, we define $\phi_{k,m}^{(n)}$ as follows.
\ba
\phi_{k,m}^{(n)}=\alpha_{m}+k \cdot \beta_{m,n}, \forall k,m \in [M], n \in [2].
\ea
Note that $\phi_{k,m}^{(n)}$ is the delay/latency for mobile user $m$ to update and upload its local model to edge server $n$ when $|A_{n}|=k$ and $m \in A_{n}$.

Let $\gamma_{k,m}^{(1)} \in [M]$ be the \textit{primary} rank of mobile user $m$ when $|A_{1}|=k$ and mobile users are sorted according to the values of $(\phi_{k,1}^{(1)},\phi_{k,2}^{(1)},..,\phi_{k,M}^{(1)})$ in \textit{decreasing} order, $\forall m \in [M]$. Specifically, if $\phi_{k,i}^{(1)} > \phi_{k,j}^{(1)}$, $\gamma_{k,i}^{(1)}<\gamma_{k,j}^{(1)}$. If $\phi_{k,i}^{(1)} = \phi_{k,j}^{(1)}$ and $i <j$, $\gamma_{k,i}^{(1)}<\gamma_{k,j}^{(1)}$. Then, $\gamma_{k,i}^{(1)} \ne \gamma_{k,j}^{(1)}$, $\forall i \ne j$. Define $\gamma_{k}^{(1)}=(\gamma_{k,1}^{(1)},\gamma_{k,2}^{(1)},..,\gamma_{k,M}^{(1)})$. For each pair $(k,r)$, where $k,r \in [M]$, let $s_{1}(k,r)$ be the unique integer in the set $[M]$ such that
\ba
\gamma_{k,s_{1}(k,r)}^{(1)} = r.
\label{Group1Leader}
\ea
Namely, $s_{1}(k,r)$ is the index of the mobile user with primary rank $r$ when $|A_{1}|=k$.

Let $h(0)$ be the HFL latency when $|A_{1}|=0$. Since $|A_{2}|=M-|A_{1}|$, we have 
\ba
h(0) &=& \max_{m: m \in [M]} \alpha_{m}+M \cdot \beta_{m,2}+d_{2}^{(2)}.
\label{h0}
\ea

Let $h(k,r)$ be the minimum HFL latency when $|A_{1}|=k$ and $\min_{m \in A_{1}} \gamma_{k,m}^{(1)}=r$, $\forall k,r \in [M]$. Namely, $h(k,r)$ is the value of the following optimization problem.
\ba
&& \min_{(A_{1},A_{2})} \max_{n \in [2]} \big[\max_{m: m \in A_{n}} d_{m}^{(1)}+d_{n}^{(2)} \big] \nonumber\\
&& \textrm{subject to} \nonumber\\
&& A_{1} \cup A_{2}=[M] \nonumber\\
&& A_{1} \cap A_{2}=\emptyset \nonumber\\
&& |A_{1}|=k \nonumber\\
&& \min_{m \in A_{1}} \gamma_{k,m}^{(1)}=r. 
\label{OptimalUserAssociationSubproblems}
\ea
The set of feasible solutions of (\ref{OptimalUserAssociationSubproblems}) is a subset of $\Omega_{2}$ with class index $(k,r)$. The above optimization problem is a subproblem of (\ref{OptimalUserAssociationTwoEdgeServers}) with index $(k,r)$. Denote an optimal solution of (\ref{OptimalUserAssociationSubproblems}) by $(\Xi_{k,r}^{(1)},\Xi_{k,r}^{(2)})$.

Let $\Theta_{k,r}$ be the set of feasible solutions of (\ref{OptimalUserAssociationSubproblems}), $\forall k,r \in [M]$. Specifically,
\ba
\Theta_{k,r} &=& \{(A_{1},A_{2})| A_{1} \cup A_{2}=[M],A_{1} \cap A_{2}=\emptyset,\nonumber\\
&& |A_{1}|=k,\min_{m \in A_{1}} \gamma_{k,m}^{(1)}=r\}.
\ea
Note that an element of $\Theta_{k,r} $ is a partition of $[M]$ and therefore $\Theta_{k,r} \subset \Omega_{2}$, $\forall k,r \in [M]$. 

Define $\Theta_{0}$ as follows. 
\ba
\Theta_{0}=\{(\emptyset,[M]) \}.
\ea
It is clear that $\Theta_{0} \in \Omega_{2}$. 

When $(A_{1},A_{2}) \in \Omega_{2}$, $|A_{1}|=0$ or $|A_{1}| \in [M]$. If $|A_{1}|=0$, $A_{2}=[M]$. On the other hand, if $|A_{1}|=k \in [M]$, $\min_{m \in A_{1}} \gamma_{k,m}^{(1)} \in [M]$. Hence, we have
\ba
\Omega_{2} &=& \Theta_{0} \cup (\cup_{(k,r): k,r \in [M]} \Theta_{k,r}).
\label{OmegaPartition}
\ea
The following theorem states that one can obtain the value of $h^{*}$ based on the values of $h(0)$ and $h(k,r)$'s.

\textbf{Theorem 1:} When $N=2$,
\ba
h^{*} &=& \min (h(0),\min_{(k,r): k,r \in [M]} h(k,r)). \nonumber
\ea

\textit{Proof:} 

1. Based on (\ref{EdgeLatency}) and (\ref{OptimalUserAssociationTwoEdgeServers}), 
\ba
h^{*} &=& \min_{(A_{1},A_{2}) \in \Omega_{2}} \max_{n: n \in [2]} \big[\max_{m: m \in A_{n}} \alpha_{m}+|A_{n}| \cdot \beta_{m,n} \nonumber\\
&& +d_{n}^{(2)} \big]. \nonumber
\ea

2. For each $(A_{1},A_{2}) \in \Omega_{2}$, define $f(A_{1},A_{2})$ as follows.
\ba
&& f(A_{1},A_{2}) \nonumber\\
&=& \max_{n: n \in [2]} \big[\max_{m: m \in A_{n}} \alpha_{m}+|A_{n}| \cdot \beta_{m,n}+d_{n}^{(2)} \big]. \nonumber
\ea

Thus,
\ba
h^{*} &=&  \min_{(A_{1},A_{2}) \in \Omega_{2}} f(A_{1},A_{2}). \nonumber
\ea

3. According to (\ref{OmegaPartition}), $\Omega_{2} = \Theta_{0} \cup (\cup_{(k,r): k,r \in [M]} \Theta_{k,r})$. Recall that $\Theta_{0}$ consists of one element. Then, we have
\ba
h^{*} &=& \min (f(\Theta_{0}), \min_{(k,r): k,r \in [M]} \min_{(A_{1},A_{2}) \in \Theta_{k,r}} f(A_{1},A_{2}) ) \nonumber\\
&=& \min (h(0),\min_{(k,r): k,r \in [M]} h(k,r)). \nonumber
\ea
The second equality is due to that $f(\Theta_{0})=h(0)$ and $h(k,r)=\min_{(A_{1},A_{2}) \in \Theta_{k,r}} f(A_{1},A_{2})$, $\forall k,r \in [M]$. \qedb

Based on Theorem 1, one can obtain the value of $h^{*}$ based on the values of $M^{2}+1$ variables. If one can obtain the value for each of the $M^{2}+1$ variables in polynomial time, one can obtain the value of $h^{*}$ in polynomial time. According to (\ref{h0}), one can obtain the value of $h(0)$ in polynomial time. 

In order to obtain the value of $h(k,r)$ in polynomial time, consider the case in which $|A_{1}|=k$ and $\min_{m \in A_{1}} \gamma_{k,m}^{(1)}=r$. In this case, if $m \in [M]$ and $\gamma_{k,m}^{(1)}=r$, mobile user $m$ has to be associated with edge server $1$. In addition, if $m \in [M]$ and $\gamma_{k,m}^{(1)} \in [r-1]$, mobile user $m$ cannot be associated with edge server $1$ and therefore has to be associated with edge server $2$. Furthermore, if $m \in [M]$ and $\gamma_{k,m}^{(1)}>r$, mobile user $m$ is associated with either edge server $1$ or edge server $2$. Therefore, we define $\Lambda_{k,r}$ as the set that consists of the indexes of mobile users each with primary rank larger than $r$. Specifically,
\ba
\Lambda_{k,r} &=& \{m \in [M]| r+1 \le \gamma_{k,m}^{(1)} \le M\}.
\label{Lambda}
\ea
It is clear that $|\Lambda_{k,r}|=M-r$. To obtain the value of $h(k,r)$, among the $M-r$ mobile users with indexes in $\Lambda_{k,r}$, one has to optimally assign $k-1$ mobile users to edge server $1$ and the remaining $(M-r)-(k-1)=M-r-k+1$ mobile users to edge server $2$. 

To optimally assign the $M-r$ mobile users with indexes in $\Lambda_{k,r}$ to the two edge servers, we rank mobile users with indexes in $\Lambda_{k,r}$ from the viewpoint of edge server $2$. as follows. First, for each $m \in \Lambda_{k,r}$, we assign $\phi_{M-k,m}^{(2)}$ to mobile user $m$. Next, we sort the mobile users with indexes in $\Lambda_{k,r}$ according to $\phi_{M-k,m}^{(2)}$'s in \textit{increasing} order. The corresponding rank for mobile user $m$ is denoted by $\gamma_{k,r,m}^{(2)}$ and is called the \textit{secondary} rank of mobile user $m$ when $|A_{1}|=k$ and $\min_{m \in A_{1}} \gamma_{k,m}^{(1)}=r$ , $\forall m \in \Lambda_{k,r}$. Specifically, $\gamma_{k,r,m}^{(2)} \in [M-r]$, $\forall m \in \Lambda_{k,r}$. In addition, if $\phi_{M-k,i}^{(2)} < \phi_{M-k,j}^{(2)}$, $\gamma_{k,r,i}^{(2)}<\gamma_{k,r,j}^{(2)}$, $\forall i,j \in \Lambda_{k,r}$. Furthermore, if $\phi_{M-k,i}^{(2)} = \phi_{M-k,j}^{(2)}$ and $i <j$, $\gamma_{k,r,i}^{(2)}<\gamma_{k,r,j}^{(2)}$, $\forall i,j \in \Lambda_{k,r}$. Then, if $i \ne j$, $\gamma_{k,r,i}^{(2)} \ne \gamma_{k,r,j}^{(2)}$, $\forall i,j \in \Lambda_{k,r}$. 

Based on (\ref{Lambda}), if $m \in \Lambda_{k,r}$, $\gamma_{k,m}^{(1)} \ge r+1$ and therefore $d_{m}^{(1)} \le d_{s_{1}(k,r)}^{(1)}$. Thus, if $s_{1}(k,r) \in A_{1}$ and $A_{1} \subset \{s_{1}(k,r)\} \cup \Lambda_{k,r}$, $\max_{m: m \in A_{1}} d_{m}^{(1)}+d_{1}^{(2)}=d_{s_{1}(k,r)}^{(1)}+d_{1}^{(2)}$. Hence, to obtain $h(k,r)$, one has to assign the $M-r-k+1$ mobile users with the highest secondary ranks among the mobile users with indexes in $\Lambda_{k,r}$ to edge server $2$. Since there are polynomial-time sorting algorithms \cite{Algorithms}, the assignment can be completed in polynomial time. 

\textbf{Theorem 2:}
\ba
\Xi_{k,r}^{(2)} &=& \{m \in [M]| \gamma_{k,m}^{(1)} \in [r-1]\} \cup \nonumber\\
&& \{m \in \Lambda_{k,r}| \gamma_{k,r,m}^{(2)} \in [M-r-k+1]\} \nonumber\\
\Xi_{k,r}^{(1)} &=& \{m \in [M]| m \notin \Xi_{k,r}^{(2)}\}. 
\label{XiSet}
\ea

\textit{Proof:} 

1. Consider the optimization problem in (\ref{OptimalUserAssociationTwoEdgeServers}). By definition, $(\Xi_{k,r}^{(1)},\Xi_{k,r}^{(2)})$ is the optimal values of $(A_{1},A_{2})$ that minimize the objective function given that $|A_{1}|=k$ and $\min_{m \in A_{1}} \gamma_{k,m}^{(1)}=r$. Consider mobile user $m$ with $\gamma_{k,m}^{(1)} \in [r-1]$. When $\min_{m \in A_{1}} \gamma_{k,m}^{(1)}=r$, mobile user $m$ cannot be assigned to edge server $1$ since $\gamma_{k,m}^{(1)} <r$ and therefore has to be assigned to edge server $2$ since $N=2$. Hence, $\{m \in [M]| \gamma_{k,m}^{(1)} \in [r-1]\}  \subseteq \Xi_{k,r}^{(2)}$.

2. Based on (\ref{Lambda}), $\Lambda_{k,r} = \{m \in [M]| r+1 \le \gamma_{k,m}^{(1)} \le M\}$. Thus, $\Lambda_{k,r}=[M] \setminus  \{m \in [M]| \gamma_{k,m}^{(1)} \in [r] \}$. When $|A_{1}|=k$, $|A_{2}|=M-k$. To obtain the value of $\Xi_{k,r}^{(2)}$, in addition to the $r-1$ elements in $\{m \in [M]| \gamma_{k,m}^{(1)} \in [r-1]\}$, one has to optimally assign $M-k-(r-1)$ elements in $\Lambda_{k,r}$ to $\Xi_{k,r}^{(2)}$.

3. If $m \in \Lambda_{k,r}$, $\gamma_{k,m}^{(1)} > r$ and therefore $\phi_{k,m}^{(1)} \le \phi_{k,s_{1}(k,r)}^{(1)}$. Thus, if $|A_{1}|=k$ and $A_{1} \subseteq \{s_{1}(k,r)\} \cup \Lambda_{k,r}$, $\max_{m \in A_{1}} \phi_{k,m}^{(1)} = \phi_{k,s_{1}(k,r)}^{(1)}$ and therefore $\max_{m \in A_{1}} d_{m}^{(1)}+d_{1}^{(2)} = \phi_{k,s_{1}(k,r)}^{(1)}+d_{1}^{(2)}$, which is a constant and does not depend on $A_{1}$ for each fixed $(k,r)$. 

4. When $|A_{1}|=k$ and $A_{1} \subseteq \{s_{1}(k,r)\} \cup \Lambda_{k,r}$, to minimize $\max_{n \in [2]} \max_{m \in A_{n}} d_{m}^{(1)}+d_{n}^{(2)}$, based on 3, it is sufficient to minimize $\max_{m \in A_{2}} d_{m}^{(1)}+d_{2}^{(2)} $. Note $\{m \in [M]| \gamma_{k,r,m}^{(2)} \in [M-k-r+1]\} \subseteq \Lambda_{k,r}$ and for each $m \in \Lambda_{k,r}$, $\gamma_{k,r,m}^{(2)}$ is obtained by sorting $\phi_{M-k,m}^{(2)}$'s in increasing order. Hence, $\Xi_{k,r}^{(2)} = \{m \in [M]| \gamma_{k,m}^{(1)} \in [r-1]\} \cup  \{m \in \Lambda_{k,r}| \gamma_{k,r,m}^{(2)} \in [M-k-r+1]\}$.

5. Since $(\Xi_{k,r}^{(1)},\Xi_{k,r}^{(2)})$ is a partition of $[M]$, $\Xi_{k,r}^{(1)} = \{m \in [M]| m \notin \Xi_{k,r}^{(2)}\}$. \qedb

For each pair $(k,r)$, where $k,r \in [M]$, after obtaining the values of $\phi_{k,m}^{(1)}$'s, $\phi_{M-k,m}^{(2)}$'s and $(\Xi_{k,r}^{(1)},\Xi_{k,r}^{(2)})$, one can obtain the value of $h(k,r)$ in polynomial time as follows. Define $\zeta_{k,r}^{(n)}$ as the latency for mobile users to upload local models to edge server $n$ when $A_{1}=\Xi_{k,r}^{(1)}$ and $A_{2}=\Xi_{k,r}^{(2)}$. Then,
\ba
\zeta_{k,r}^{(1)} &=&  \max_{m: m \in \Xi_{k,r}^{(1)} } \phi_{k,m}^{(1)} \nonumber\\
&=& \phi_{k,s_{1}(k,r)}^{(1)}. 
\label{zeta1}
\ea

In addition,
\ba
\zeta_{k,r}^{(2)} &=& \max_{m: m \in \Xi_{k,r}^{(2)} } \phi_{M-k,m}^{(2)}.
\label{zeta2}
\ea

Furthermore,
\ba
h(k,r) &=& \max_{n: n \in [2]} \zeta_{k,r}^{(n)}+d_{n}^{(2)}.
\ea

\textbf{Theorem 3:} The twin sorting dynamic programming algorithm obtains an optimal solution for (\ref{OptimalUserAssociationVersion1}) when $N=2$.

\textit{Proof:}

1. When $N=2$, (\ref{OptimalUserAssociationVersion1}) is equivalent to (\ref{OptimalUserAssociationTwoEdgeServers}). In line 1 of Algorithm \ref{Algorithm-TSDP}, the TSDP algorithm obtains $h(0)$. In addition, the TSDP algorithm obtains $(\Xi_{k,r}^{(1)},\Xi_{k,r}^{(2)})$ in line 12 based on Theorem 2 and $h(k,r)$ in line 14 in Algorithm \ref{Algorithm-TSDP}, $\forall k,r \in [M]$.

2. Based on Theorem 1, $h^{*} = \min (h(0),\min_{(k,r): k,r \in [M]} h(k,r))$. The TSDP algorithm updates the value of $h^{*}$ and $(A_{1}^{*},A_{2}^{*})$ in lines 15-17 In Algorithm \ref{Algorithm-TSDP}. Thus, after comparing $h(0)$ and $h(k,r)$'s, the TSDP algorithm obtains the value of $h^{*}$ and $(A_{1}^{*},A_{2}^{*})$ when it terminates. \qedb

Let $\Lambda_{k,r}[\ell]$ be the $\ell$th element in the set $\Lambda_{k,r}$, $\forall \ell$. We define two vectors that are used in Algorithm \ref{Algorithm-TSDP} as follows. First,
\ba
u_{k,r}=(\phi_{M-k,\Lambda_{k,r}[1]}^{(2)},\phi_{M-k,\Lambda_{k,r}[2]}^{(2)},..,\phi_{M-k,\Lambda_{k,r}[M-r]}^{(2)}).
\label{uVector}
\ea
In addition, 
\ba
\gamma_{k,r}^{(2)}=(\gamma_{k,r,\Lambda_{k,r}[1]}^{(2)},\gamma_{k,r,\Lambda_{k,r}[2]}^{(2)},..,\gamma_{k,r,\Lambda_{k,r}[M-r]}^{(2)}).
\ea
The secondary rank vector $\gamma_{k,r}^{(2)}$ consists of the secondary ranks of mobile users with indexes in $\Lambda_{k,r}$. One can obtain the vector $\gamma_{k,r}^{(2)}$ by sorting elements in the vector $u_{k,r}$.

\textbf{Theorem 4:} The computational complexity of the proposed TSDP algorithm is $O(M^{3}\log_{2}M)$. 

\textit{Proof:}

1. Consider Algorithm \ref{Algorithm-TSDP}. In line 1, it takes $O(M)$ time to obtain $h^{*}$. For each $k \in [M]$, it takes $O(M)$ to obtain $\phi_{k,m}^{(1)}$'s in line 4 and $O(M \log_{2}M)$ time to acquire $\gamma_{k}^{(1)}$ in line 5. Given $\gamma_{k}^{(1)}$, for each $r \in [M]$, it takes $O(1)$ time to obtain $s_{1}(k,r)$ in line 7 and $O(M)$ time to get $\Lambda_{k,r}$ in line 8. In addition, it takes $O(M)$ time to run the codes in lines 9-10. In line 11, it takes $O(M \log_{2} M)$ time to acquire $\gamma_{k,r}^{(2)}$ based on QuickSort \cite{Algorithms}. 

2. Based on Theorem 2, it takes $O(M)$ time to obtain $(\Xi_{k,r}^{(1)},\Xi_{k,r}^{(2)})$ in line 12. In line 13, the algorithm spends $O(M)$ time for obtaining $(\zeta_{k,r}^{(1)},\zeta_{k,r}^{(2)})$. In line 14, it takes $O(1)$ time to obtain $h(k,r)$. The computational complexity of codes in line 16 is $O(M)$. Thus, for each $r \in [M]$, the overall computational complexity of the codes in lines 7-17 is equal to $O(1)+O(M)+O(M)+O(M \log_{2} M)+O(M)+O(M)+O(1)+O(M)=O(M \log_{2} M)$. 

3. Hence, the aggregated computational complexity of the codes in lines 3-19 is equal to $M \cdot [O(M \log_{2} M)+M \cdot O(M \log_{2} M)]=O(M^{3}\log_{2}M)$. For each $m \in [M]$, it takes $O(M)$ time to check if $m \in A_{1}^{*}$. Thus, the computational complexity of codes in lines 20-26 is $M \cdot O(M)=O(M^{2})$. Therefore, the computational complexity of Algorithm \ref{Algorithm-TSDP} is equal to $O(M^{3}\log_{2}M)+O(M^{2})=O(M^{3}\log_{2}M)$. \qedb

Let $s \in [M]$ be an integer. With minor modifications, the TSDP algorithm could solve the following combinatorial optimization problem in polynomial time and select $s$ mobile users to minimize the HFL latency.
\ba
&& \min_{(A_{1},A_{2})} \max_{n \in [2]} \big[\max_{m: m \in A_{n}} d_{m}^{(1)}+d_{n}^{(2)} \big] \nonumber\\
&& \textrm{subject to} \nonumber\\
&& A_{1} \cup A_{2} \subseteq [M] \nonumber\\
&& |A_{1} \cup A_{2}|=s \nonumber\\
&& A_{1} \cap A_{2}=\emptyset.
\label{OptimalUserSelection}
\ea
We omit the details due to the limit of space.

\begin{algorithm}
\captionsetup{labelsep=period}
\caption{\texttt{TSDP-assisted}}
\label{Algorithm-TSDP-assisted}
    \begin{algorithmic}[1]
    \Require $M$, $N$, $\alpha$, $\beta$, $d^{(2)}$, $J$. 
    \Ensure $\mathbf{X}^{\dag}$, $\Theta^{\dag}$, $h^{\dag}$, $A^{\dag}:=(A_{1}^{\dag},A_{2}^{\dag},..,A_{N}^{\dag})$.
    \State // 1. Use a baseline algorithm to obtain an initial solution.
    \State $(h^{\dag},A^{\dag})=\texttt{Max-SNR}(M,N,\alpha,\beta,d^{(2)})$.
    \State // 2. For each pair of adjacent edge servers, use the TSDP algorithm to obtain a better solution.
    \State $h^{\dag} \gets 0$.
    \For{$k=1$ to $\lfloor \frac{N}{2} \rfloor$}
    	\State $(u,v) \gets (2k-1,2k)$.
	\State $\Psi \gets A^{*}_{u} \cup A^{*}_{v}$, $\tilde{M} \gets |\Psi|$.
	\State $\tilde{d}^{(2)}_{1} \gets d^{(2)}_{u}$, $\tilde{d}^{(2)}_{2} \gets d^{(2)}_{v}$, $i \gets 1$.
	\ForEach{$m \in \Psi$}
		\State $\tilde{\alpha}_{i} \gets \alpha_{m}$.
		\State $\tilde{\beta}_{i,1} \gets \beta_{m,u}$, $\tilde{\beta}_{i,2} \gets \beta_{m,v}$.
		\State $i \gets i+1$.
	\EndFor
	\State $(h,A_{1},A_{2})=\texttt{TSDP}(\tilde{M},\tilde{\alpha},\tilde{\beta},\tilde{d}^{(2)})$.
	\State $h^{\dag} \gets \max(h^{\dag},h)$, $(A^{\dag}_{u},A^{\dag}_{v}) \gets (A_{1},A_{2})$.	
     \EndFor
     \State // Check if $N$ is an odd number.
     \If{$N>2 \lfloor \frac{N}{2} \rfloor$}
     	\State $h \gets \max_{m: m \in A^{\dag}_{N}} \alpha_{m}+|A^{\dag}_{N}| \cdot \beta_{m,N}+d_{N}^{(2)}$.
     	\State $h^{\dag} \gets \max(h^{\dag},h)$.
     \EndIf
    \State // 3. Use a greedy algorithm to improve user association.
    \State $\mathbf{Y}_{0} \gets $\texttt{Partition2Matrix}$(A^{\dag}_{1},A^{\dag}_{2},..,A^{\dag}_{N})$. 
    \For{$m=1$ to $M$}
    	\State $\mathbf{Y}_{m} \gets \texttt{GST}(m,\mathbf{Y}_{m-1},M,N,\alpha,\beta,d^{(2)})$.
    \EndFor
    \State $\mathbf{X}_{3} \gets \mathbf{Y}_{M}$.
    \State // 4. Use dynamic bandwidth allocation.
    \State $(A_{1}^{\dag},A_{2}^{\dag},..,A_{n}^{\dag}) \gets \texttt{Matrix2Partition}(\mathbf{X}_{3})$.
    \For{$n=1$ to $N$}
    	\State Obtain $(\mu_{n}^{*},\theta_{n}^{*})$ by solving (\ref{OptimalDBA2}). 
    \EndFor
    \State $h^{\dag} \gets \max_{n \in [N]} \mu_{n}^{*}+d_{n}^{(2)}$.		
    \State $A^{\dag} \gets (A_{1}^{\dag},A_{2}^{\dag},..,A_{n}^{\dag})$, $\Theta^{\dag} \gets [\theta_{1}^{*},\theta_{2}^{*},..,\theta_{n}^{*}]$.
    \State // 5. Use a CPR algorithm.
    \For{$k=1$ to $10$}
    	\State $(h^{\dag},A^{\dag},\Theta^{\dag}) \gets$\texttt{CPR-M}$(M,N,\alpha,\beta,d^{(2)},h^{\dag},A^{\dag},\Theta^{\dag},J)$.
    \EndFor
    \end{algorithmic}
\end{algorithm}    

\section{The TSDP-assisted algorithm}

In this section, we put forward the TSDP-assisted algorithm for determining user association and bandwidth allocation in an HFL round when $N \ge 3$. Consider a round of HFL. Let $\mathbf{X}^{\dag}$ be the user association matrix produced by the TSDP-assisted algorithm. Let $\Theta^{\dag}$ be the bandwidth allocation matrix produced by the TSDP-assisted algorithm such that $[\Theta^{\dag}]_{m,n}$ is the fraction of bandwidth that edge server $n$ allocates to mobile user $m$, $\forall m \in [M], n \in [N]$. Pseudo codes for the TSDP-assisted algorithm are included in Algorithm \ref{Algorithm-TSDP-assisted}.

The TSDP-assisted algorithm consists of five phases. In the first phase, the TSDP-assisted algorithm utilizes a baseline algorithm such as Max-SNR to obtain an initial solution for user association when each edge sever adopts equal bandwidth allocation. In the second phase, it adopts the TSDP algorithm to find a better user association matrix for each pair of edge servers. In the third phase, it uses the greedy server transfer (GST) algorithm. In the fourth phase, it makes use of dynamic bandwidth allocation rather than equal bandwidth allocation. In the fifth phase, it exploits the critical path reduction (CPR) algorithm. 

We now elaborate on the technical details of the proposed TSDP-assisted algorithm. In the second phase, the TSDP-assisted algorithm treats edge server $2k-1$ and edge server $2k$ as the $k$th pair of edge servers, $\forall 1 \le k \le \lfloor \frac{N}{2} \rfloor$. Next, the TSDP-assisted algorithm uses $\lfloor \frac{N}{2} \rfloor$ rounds to update the user association matrix. Specifically, in the $k$th round, it utilizes the TSDP algorithm to find an optimal user association matrix for mobile users that are associated with the $k$th pair of edge servers in phase 1, $\forall 1 \le k \le \lfloor \frac{N}{2} \rfloor$. Let $\mathbf{X}_{2}$ be the user association matrix produced by the TSDP-assisted algorithm in the second phase.

In the third phase, given $\mathbf{X}_{2}$, the TSDP-assisted algorithm uses the proposed GST algorithm $M$ times to find a better user association matrix. Specifically, the GST algorithm is a greedy algorithm that contains $M$ rounds and discovers an optimal edge server for mobile user $m$ to connect in round $m$, $\forall m \in [M]$. We now introduce some variables used in the GST algorithm. Define $\mathbf{Y}_{0}=\mathbf{X}_{2}$. In addition, let $\mathbf{Y}_{m} \in \{0,1\}^{M \times N}$ be the user association matrix in the end of round $m$ of the third phase, $\forall m \in [M]$. For each $\mathbf{Y} \in \{0,1\}^{M \times N}$, let $\Delta_{m,n}(\mathbf{Y}) \in \{0,1\}^{M \times N}$ be a matrix such that $[\Delta_{m,n}(\mathbf{Y})]_{m,n}=1$ and $[\Delta_{m,n}(\mathbf{Y})]_{m,n'}=0$, $\forall n' \ne n$. Let $\ell_{m,n}$ be the HFL latency when the user association matrix $\mathbf{X}$ is equal to $\Delta_{m,n}(\mathbf{Y}_{m-1})$, $\forall m \in [M], n \in [N]$. Define $k^{*}(m)=\arg \min_{n \in [N]} \ell_{m,n}$, $\forall m \in [M]$. According to the GST algorithm, $[\mathbf{Y}_{m}]_{i,n}=[\mathbf{Y}_{m-1}]_{i,n}$, $\forall i \ne m, n \in [N]$. In addition, $[\mathbf{Y}_{m}]_{m,n}=1$ if $n=k^{*}(m)$ and $[\mathbf{Y}_{m}]_{m,k}=0$ if $n \ne k^{*}(m)$. Let $\mathbf{X}_{3}$ be the user association matrix produced by the TSDP-assisted algorithm in the third phase. Based on the TSDP-assisted algorithm, $\mathbf{X}_{3}=\mathbf{Y}_{M}$. Pseudo codes for the GST algorithm are included in Algorithm \ref{Algorithm-GST}. 


In the fourth phase, the TSDP-assisted algorithm adopts dynamic bandwidth allocation (DBA). Recall that $A_{n}(t)$ is the set that consists of the indexes of mobile users that are associated with edge server $n$ in HFL round $t$. In addition, $\theta_{m,n}(t)$ is the fraction of bandwidth that edge server $n$ assigns to mobile user $m$ in HFL round $t$. Let $\theta_{n}(t)=(\theta_{1,n}(t),\theta_{2,n}(t),..,\theta_{M,n}(t))^{T}$, $\forall n \in [N], t \in \mathbb{N}$. Moreover, define $\Theta(t) \in \mathbb{R}^{M \times N}$ as a matrix such that the element in the $i$th row and $j$th column is equal to $\theta_{i,j}(t)$. To obtain an optimal bandwidth allocation vector at edge server $n$ in HFL round $t$, we formulate the subsequent optimization problem.
\ba
&& \min_{\theta_{n}(t) \in \mathbb{R}^{M}} \max_{m: m \in A_{n}(t)} \alpha_{m}(t)+\frac{\beta_{m,n}(t)}{\theta_{m,n}(t)} \nonumber\\
&& \textrm{subject to} \nonumber\\
&& \theta_{m,n}(t) \in [0,1], \forall m \in [M] \nonumber\\
&& \sum_{m=1}^{M} \theta_{m,n}(t) \le 1 \nonumber\\
&& \theta_{m,n}(t)=0, \forall m \notin A_{n}(t).
\label{OptimalDBA1}
\ea 
The value of $\max_{m: m \in A_{n}(t)} \alpha_{m}(t)+\frac{\beta_{m,n}(t)}{\theta_{m,n}(t)}$ is the maximum latency for all mobile users associated with edge server $n$ in HFL round $t$ to update and transmit their local models to edge server $n$ in HFL round $t$. 

After defining $\mu_{n}(t)=\max_{m: m \in A_{n}(t)} \alpha_{m}(t)+\frac{\beta_{m,n}(t)}{\theta_{m,n}(t)}$, we transform (\ref{OptimalDBA1}) into the following optimization problem. 
\ba
&& \min_{(\theta_{n}(t),\mu_{n}(t)) \in \mathbb{R}^{M} \times \mathbb{R}} \mu_{n}(t) \nonumber\\
&& \textrm{subject to} \nonumber\\
&&  \mu_{n}(t) \in [0,\infty) \nonumber\\
&& \mu_{n}(t) \ge \alpha_{m}(t)+\frac{\beta_{m,n}(t)}{\theta_{m,n}(t)}, \forall m \in A_{n}(t) \nonumber\\
&& \theta_{m,n}(t) \in [0,1], \forall m \in [M] \nonumber\\
&& \sum_{m=1}^{M} \theta_{m,n}(t) \le 1 \nonumber\\
&& \theta_{m,n}(t)=0, \forall m \notin A_{n}(t).
\label{OptimalDBA2}
\ea 

\begin{algorithm}
\captionsetup{labelsep=period}
\caption{\texttt{GST}: greedy server transfer}
\label{Algorithm-GST}
    \begin{algorithmic}[1]
    \Require $m$, $\mathbf{Y}_{m-1}$, $M$, $N$, $\alpha$, $\beta$, $d^{(2)}$. 
    \Ensure $\mathbf{Y}_{m}$.
    \State $s \gets \arg \max_{n: n \in [N]} x_{m,n}$, $d \gets s$.
    \For{$k=1$ to $N$}
    	\State $\mathbf{X} \gets \Delta_{m,k}(\mathbf{Y})$.
	\For{$n=1$ to $N$}
		\State $A_{n} \gets \{m \in [M] | [\mathbf{X}]_{m,n}=1\}$.
	\EndFor
	\State $\ell_{m,k} \gets \max_{n \in [N]} \max_{m \in A_{n}} \alpha_{m}+|A_{n}| \cdot \beta_{m,n}+d_{n}^{(2)}$.
    \EndFor	
    \State $k^{*} \gets \arg \min_{n \in {N}} \ell_{m,n}$.
    \State $\mathbf{Y}_{m} \gets \mathbf{Y}_{m-1}$.
    \For{$n=1$ to $N$}
    	\If{$n=k^{*}$}
		\State $[\mathbf{Y}_{m}]_{m,n} \gets 1$.	
	\Else
		\State $[\mathbf{Y}_{m}]_{m,n} \gets 0$.	
	\EndIf
    \EndFor
    \end{algorithmic}
\end{algorithm}    
 
Let $\Omega_{\theta,n,t}$ be the set of feasible solutions of (\ref{OptimalDBA2}). We now prove that (\ref{OptimalDBA2}) is a convex optimization problem. First, it is clear that the objective function is a linear function of $(\theta_{n}(t),\mu_{n}(t))$. Second, since $\frac{d^{2}}{dx^{2}} [\frac{1}{x}]=\frac{2}{x^{3}}>0$, $\forall x>0$, $\alpha_{m}(t)+\frac{\beta_{m,n}(t)}{\theta_{m,n}(t)}$ is a convex function of $\theta_{m,n}(t)$. Third, since $\alpha_{m}(t)+\frac{\beta_{m,n}(t)}{\theta_{m,n}(t)}-\mu_{n}(t)$ is a convex function of $(\theta_{n}(t),\mu_{n}(t))$, $\{(\theta_{n}(t),\mu_{n}(t)) | \alpha_{m}(t)+\frac{\beta_{m,n}(t)}{\theta_{m,n}(t)}-\mu_{n}(t) \le 0\}$ is a convex set. Fourth, $\{\theta_{n}(t)| \theta_{m,n}(t) \in [0,1], \forall m \in [M], \sum_{m=1}^{M} \theta_{m,n}(t) \le 1\}$ is a convex set. Fifth, $\{\theta_{n}(t)| \theta_{m,n}(t)=0, \forall m \notin A_{n}(t)\}$ is a convex set. Last, it is clear that $\{(\theta_{n}(t),\mu_{n}(t)) \in \mathbb{R}^{M} \times \mathbb{R}| \mu_{n}(t) \in [0,\infty)\}$ is a convex set. Since $\Omega_{\theta,n,t}$ is the intersection of the above four convex sets, $\Omega_{\theta,n,t}$ is a convex set. Since the objective function is a convex function and the set of feasible solutions is convex, (\ref{OptimalDBA2}) is a convex optimization problem. Let $(\theta_{n}^{*}(t),\mu_{n}^{*}(t))=(\theta_{1,n}^{*}(t),\theta_{2,n}^{*}(t),..,\theta_{M,n}^{*}(t),\mu_{n}^{*}(t))$ be an optimal solution of (\ref{OptimalDBA2}). 

 
In the fifth phase, the TSDP-assisted algorithm uses critical path reduction (CPR). A path from mobile user $\tilde{m}$ to the cloud server through edge server $\tilde{n}$ is said to be a critical path if $\alpha_{\tilde{m}}+\frac{\beta_{\tilde{m},\tilde{n}}}{\theta_{\tilde{m},\tilde{n}}}+d_{\tilde{n}}^{(2)}=\max_{(m,n): m \in [M],n \in [N]} \alpha_{m}+I(m \in A_{n}) \times \big[\frac{\beta_{m,n}}{\theta_{m,n}}+d_{n}^{(2)} \big]$. Namely, the latency of a critical path is equal to the HFL latency. A CPR algorithm first identifies the critical edge server in a critical path and then adopts user migration or user swap to reduce the HFL latency. We put forward two CPR algorithms. The first CPR algorithm is based on user migration and is called CPR-M. Specifically, the CPR-M algorithm tries to find an edge server $n$ such that $n \ne \tilde{n}$ and the HFL latency decreases if mobile user $\tilde{m}$ becomes associated with edge server $n$. Pseudo codes for the CPR-M algorithm are included in Algorithm \ref{Algorithm-CPR-M}. Let $\mathbf{X}_{4}$ be the user association matrix in the end of phase 4 of the TSDP-assisted algorithm. The second CPR algorithm adopts user swap and is called CPR-S. Specifically, the CPR-S algorithm finds an optimal mobile user $m'$ such that the HFL latency is minimized if mobile user $\tilde{m}$ becomes associated with edge server $\arg_{n \in [N]} [\mathbf{X}_{4}]_{m',n}$ and mobile user $m'$ becomes associated with edge server $\tilde{n}$. 

\begin{algorithm}
\captionsetup{labelsep=period}
\caption{\texttt{CPR-M}: Critical Path Reduction by Migration}
\label{Algorithm-CPR-M}
    \begin{algorithmic}[1]
    \Require $M$, $N$, $\alpha$, $\beta$, $d^{(2)}$, $h^{\dag}$, $(A_{1}^{\dag},A_{2}^{\dag},..,A_{N}^{\dag})$, $\Theta^{\dag}$, $J$.
    \Ensure $h^{\dag}$, $(A_{1}^{\dag},A_{2}^{\dag},..,A_{N}^{\dag})$, $\Theta^{\dag}$.
    \State // 1. Identify the critical path from MU $\tilde{m}$ via ES $\tilde{n}$ to cloud.  
    \State $(\tilde{m},\tilde{n}) \gets \arg \max_{(m,n): m \in [M],n \in [N]} \alpha_{m}+I(m \in A_{n}^{\dag}) \times \big[\frac{\beta_{m,n}}{\theta_{m,n}^{\dag}}+d_{n}^{(2)} \big]$.
    \State // 2. Try to find a better value for $A_{\tilde{n}}$ by user-migration. 
    \State $\mathrm{Stop} \gets 0$, $n \gets 1$.
    \While{$\mathrm{Stop}=0$ and $n \le N$}
    	\If{$n \ne \tilde{n}$}
		\State $(A_{1},A_{2},..,A_{N}) \gets (A_{1}^{\dag},A_{2}^{\dag},..,A_{N}^{\dag})$. 
		\State $A_{\tilde{n}} \gets A_{\tilde{n}} \setminus \{m\}$, $A_{n} \gets A_{n} \cup \{m\}$.
		\State $\Theta \gets \Theta^{\dag}$.
		\State $\theta_{\tilde{n}} \gets \texttt{MLBS}(M,N,\tilde{n},A_{\tilde{n}},\alpha,\beta,J)$.
		\State $\theta_{n} \gets \texttt{MLBS}(M,N,n,A_{n},\alpha,\beta,J)$.
		\State $h \gets \max_{m \in [M]} \alpha_{m}+\sum_{n=1}^{N} I(m \in A_{n}) \cdot \big[ \frac{\beta_{m,n}}{\theta_{m,n}}+d_{n}^{(2)} \big]$.
		\If{$h < h^{\dag}$}
			\State $h^{\dag} \gets h$, $(A_{1}^{\dag},A_{2}^{\dag},..,A_{N}^{\dag}) \gets (A_{1},A_{2},..,A_{N})$.
			\State $\Theta^{\dag} \gets \Theta$, $\mathrm{Stop} \gets 1$.
		\EndIf
	\EndIf
	\State $n \gets n+1$.
    \EndWhile
    \end{algorithmic}
\end{algorithm}    
 
\section{Simulation Setup and Results}

In this section, we include simulation setup and results. We wrote Python programs to obtain simulation results and adopted CVXPY \cite{CVXPY} to solve convex optimization problems. We evaluate six algorithms. The first algorithm is the Max-SNR algorithm and the second algorithm is the backbone-aware greedy (BAG) algorithm \cite{Gau2024VTC}. The BAG algorithm consists of $M$ rounds. Specifically, it assigns mobile user $k$ to the optimal edge server for minimizing the HFL latency when only the first $k$ mobile users upload their local models to edge servers, $\forall k \in [M]$. The third algorithm is the proposed TSDP algorithm and the fourth algorithm is the exhaustive search algorithm. The fifth algorithm is based on the Balanced Clustering algorithm in FedCH \cite{Wang2023TMC}. It first uses the Balanced Clustering algorithm to form clusters of mobile users and then utilizes minimum-weight bipartite matching \cite{Algorithms} to assign clusters to edge servers based on the communication latencies between cluster heads and edge servers. The sixth algorithm is the proposed TSDP-assisted algorithm.

We utilize PyTorch \cite{PyTorch} to study the impacts of user association on the testing accuracy of image recognition in a hierarchical federated learning system. For proof of concept, we adopt the CIFAR-10 \cite{CIFAR-10} dataset for training and testing machine learning models. The CIFAR-10 dataset consists of $10$ classes of images. Each class contains 5,000 training images and 1,000 testing images. We adopt a convolutional neural network (CNN) that is composed of $10$ layers including $1$ input layer, $3$ convolutional layers, $2$ pooling layers, $1$ flatten layer and $3$ fully connected layers. The CNN has 591,066 trainable parameters. While different datasets typically lead to different number of rounds required for convergence, we focus on minimizing the length of each round via optimal user association in this paper. Client selection based on heterogeneous data distributions is beyond the scope of this paper.  

We first study the case in which $N=2$ and $M=4K$, where $K$ is a positive integer. The coordinates of the first edge server are $(0,0)$, while the coordinates of the second edge server are $(5,0)$. The first $2K$ mobile devices are located around $(1,0)$ and the other $2K$ mobile devices are located around $(3,0)$. In this section, for each $t \in \mathbb{N}$, $\beta_{m,1}(t)=1$, $\forall 1 \le m \le 2K$ and $\beta_{m,1}(t)=9$, $\forall 2K+1 \le m \le 4K$. Furthermore, for each $t \in \mathbb{N}$, $\beta_{m,2}(t)=16$, $\forall 1 \le m \le 2K$ and $\beta_{m,2}(t)=4$, $\forall 2K+1 \le m \le 4K$. Mobile devices are classified into two types based on their computation capabilities. Specifically, for each $t \in \mathbb{N}$, $\alpha_{m}(t)=10$, $\forall m \in \{1,2,..,K\} \cup \{2K+1,2K+2,..,3K\}$ and $\alpha_{m}(t)=20$, $\forall m \in \{K+1,K+2,..,2K\} \cup \{3K+1,3K+2,..,4K\}$. Moreover, $d_{1}^{(2)}(t)=10$ and $d_{2}^{(2)}(t) \in [10,200]$, $\forall t \in \mathbb{N}$. 

\begin{figure}
    \centering
    \includegraphics[width=7cm]{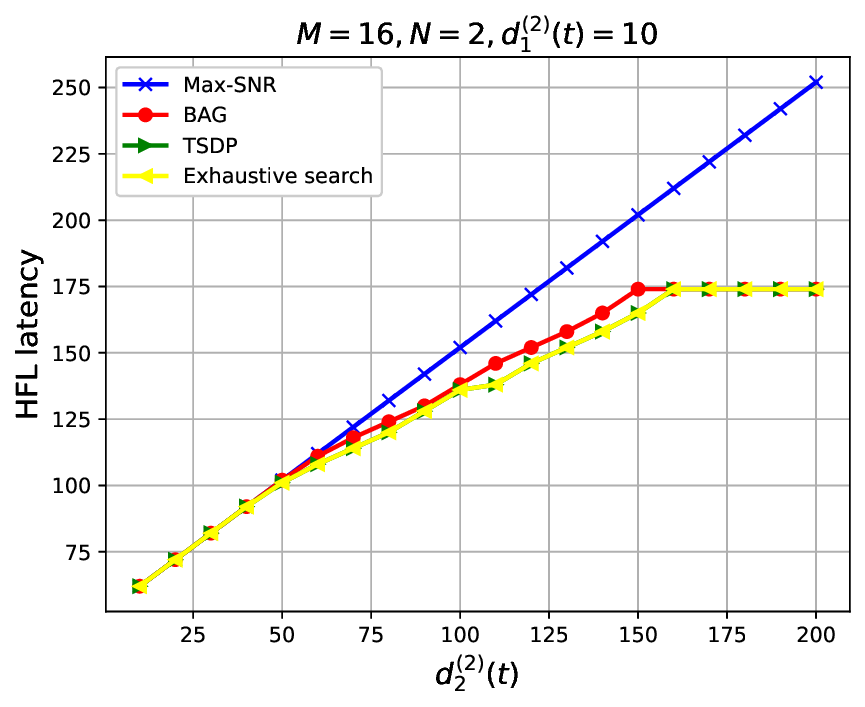}
    \caption{Advantages of the TSDP algorithm, $(M,N)=(16,2)$.}
    \label{HFL_latency_M16_N2}
\end{figure}

In Fig. \ref{HFL_latency_M16_N2}, we show the impacts of $d_{2}^{(2)}(t)$ on the HFL latency, when $M=16$ and $N=2$. For each of the four studied algorithms, the HFL latency increases as the value of $d_{2}^{(2)}(t)$ increases. It is due to that the HFL latency depends on the mobile-to-edge delays and the edge-to-cloud delays. Regardless of the value of $d_{2}^{(2)}(t)$, the HFL latency of the TSDP algorithm is identical to that of the exhaustive search algorithm. It is due to that the proposed TSDP algorithm obtains an optimal matrix of user association when $N=2$. When $d_{2}^{(2)}(t) \ge 50$, the proposed TSDP algorithm is superior to the Max-SNR algorithm. The performance improvement could be as large as $30.95\%$. When $d_{2}^{(2)}(t) \in [60,150]$, the proposed TSDP algorithm outperforms the BAG algorithm. 

\begin{figure}
    \centering
    \includegraphics[width=7cm]{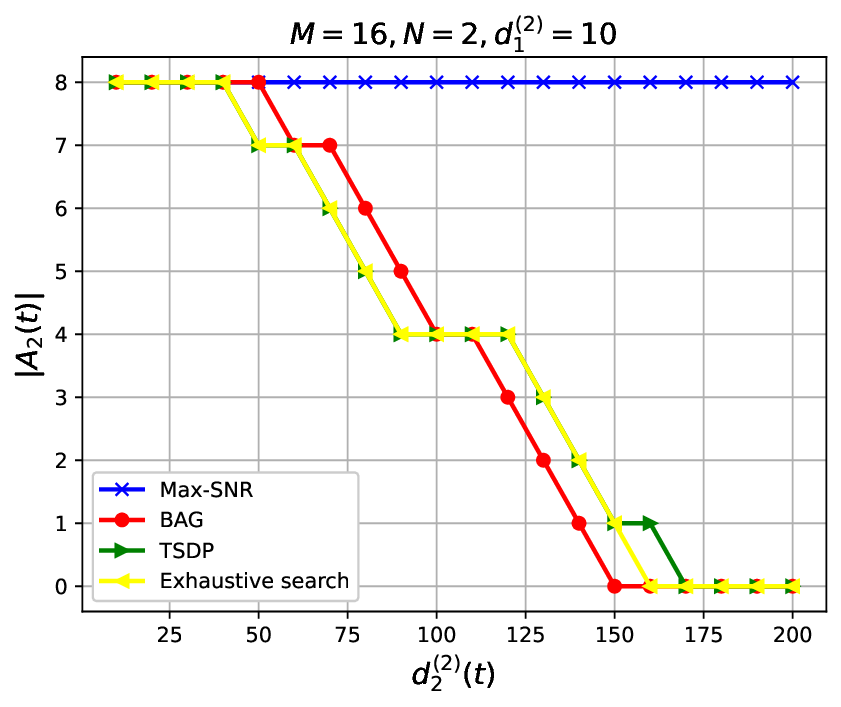}
    \caption{$|A_{2}(t)|$ for four algorithms, $(M,N)=(16,2)$.}
    \label{HFL_loading2_M16_N2}
\end{figure}

In Fig. \ref{HFL_loading2_M16_N2}, we show the impacts of $d_{2}^{(2)}(t)$ on $|A_{2}(t)|$, when $M=16$ and $N=2$. The value of $|A_{2}(t)|$ for the TSDP algorithm is almost always the same as that for the exhaustive search algorithm. It is mainly due to that each of the two algorithms obtains an optimal matrix of user association when $N=2$. However, sometimes, there are two or more optimal matrices of user association and the two algorithms obtain different optimal solutions. For the Max-SNR algorithm, the value of $|A_{2}(t)|$ remains unchanged when $d_{2}^{(2)}(t)$ changes. The Max-SNR algorithm ignores edge-to-cloud delays when assigning mobile users to edge servers. When $d_{2}^{(2)}(t) \in [60,150]$, the value of $|A_{2}(t)|$ for the BAG algorithm is different from that for the TSDP algorithm. In this case, the TSDP algorithm produces an optimal solution for user association but the BAG algorithm does not.

\begin{figure}
    \centering
    \includegraphics[width=7cm]{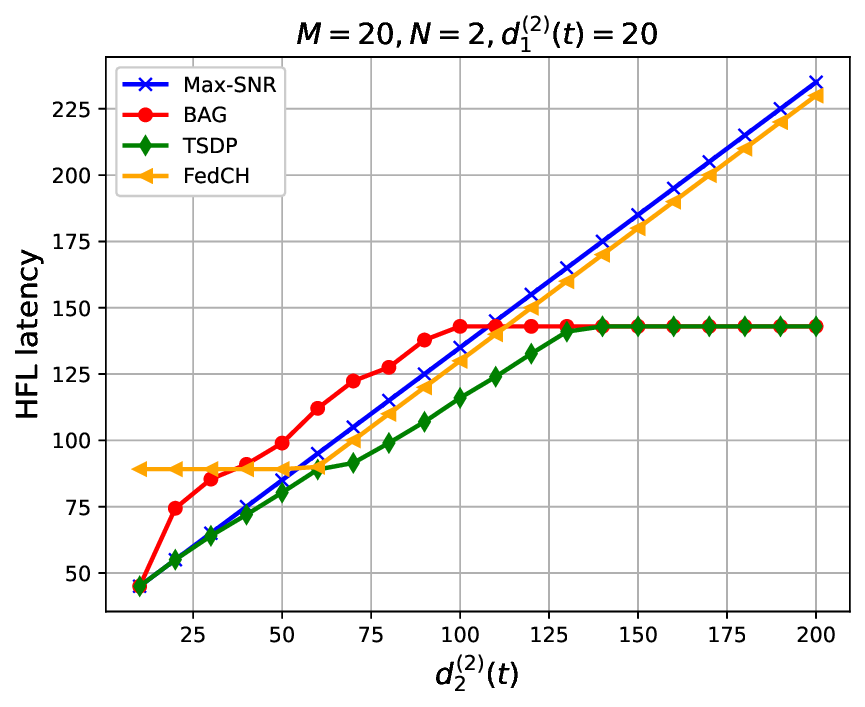}
    \caption{The HFL latency for four algorithms, $(M,N)=(20,2)$.}
    \label{HFL_latency_M20_N2_FedCH}
\end{figure}

For proof of concept, we use the following network topology to evaluate the FedCH algorithm. First, $M=20$, $N=2$, $K=\frac{M}{4}=5$, $\frac{3M}{4}$ mobile users are uniformly distributed in the line segment from $(9.5,0)$ to $(10.5,0)$ and $\frac{M}{4}$ mobile users are uniformly distributed in the line segment from $(-10.5,0)$ to $(-9.5,0)$. The first edge server is at $(-10,10)$ and the second edge server is at $(10,10)$. For each $t \in \mathbb{N}$, $\alpha_{m}(t)=10$, $\forall m \in \{1,2,..,K\} \cup \{2K+1,2K+2,..,3K\}$ and $\alpha_{m}(t)=20$, $\forall m \in \{K+1,K+2,..,2K\} \cup \{3K+1,3K+2,..,4K\}$. Let $\mathbf{u}_{m} \in \mathbb{R}^{2}$ be the coordinates of mobile user $m$ and $\mathbf{s}_{n} \in \mathbb{R}^{2}$ be the coordinates of edge server $m$, $\forall m \in [M], n \in [N]$. Let $\rho$ be a positive real number. For each $t \in \mathbb{N}$, $\beta_{m,n}(t)=\rho \cdot \norm{\mathbf{u}_{m}-\mathbf{s}_{n}}^{2}$, $\forall m \in [M], n \in [N]$.

In Fig. \ref{HFL_latency_M20_N2_FedCH}, we compare the performance of four algorithms when $\rho=0.01$. Regardless of the value of $d_{2}^{(2)}(t)$, the proposed TSDP algorithm achieves the minimum HFL latency among the four studied algorithms. When $d_{2}^{(2)}(t) \in [10,50]$, the Max-SNR algorithm outperforms the BAG algorithm and the FedCH algorithm. When $d_{2}^{(2)}(t) \in [60,110]$, the FedCH algorithm is superior to the Max-SNR algorithm and the BAG algorithm. When $d_{2}^{(2)}(t) \in [120,200]$, the BAG algorithm outperforms the Max-SNR algorithm and the FedCH algorithm. Furthermore, the Max-SNR algorithm, the BAG algorithm and the FedCH algorithm sometimes but not always are as good as the TSDP algorithm. The FedCH algorithm seeks to partition mobile users into clusters of equal size and indirectly reduces the HFL latency by assigning adjacent mobile users that have similar computing capabilities to the same cluster. The proposed TSDP algorithm directly minimizes the HFL latency by solving a challenging combinatorial optimization problem and finding an optimal user association matrix.    

\begin{figure}
    \centering
    \includegraphics[width=7cm]{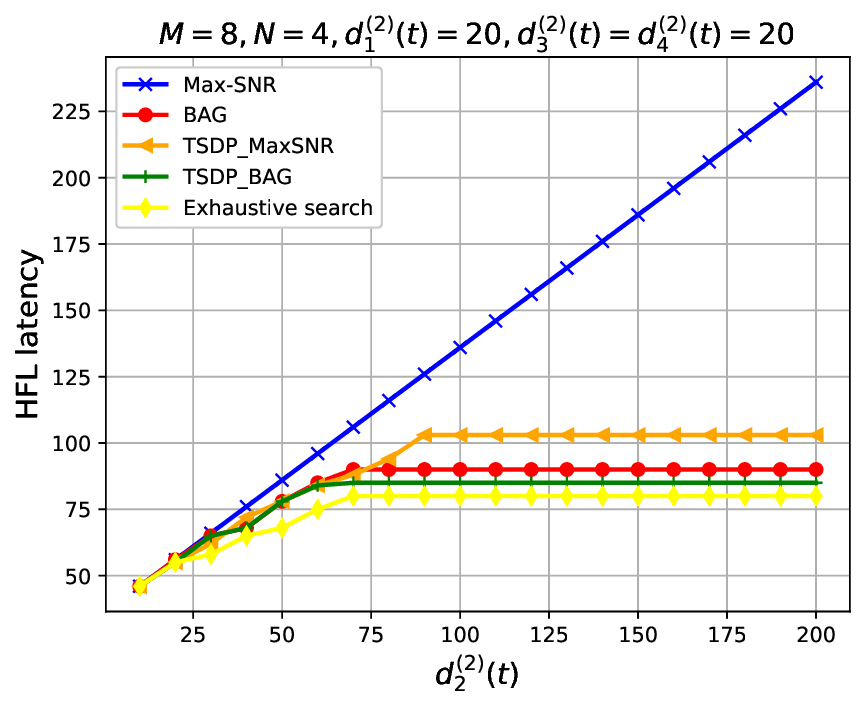}
    \caption{Advantages of the TSDP-assisted algorithm when $(M,N)=(8,4)$.}
    \label{HFL_latency_M8_N4}
\end{figure}

We evaluate the Max-SNR algorithm, the BAG algorithm, the exhaustive search algorithm and the TSDP-assisted algorithm when $N=4$. In this subsection, for the TSDP-assisted algorithm, only the first three phases are activated. When the TSDP-assisted algorithm adopts the Max-SNR algorithm to obtain an initial solution, it is called TSDP-MaxSNR. When the TSDP-assisted algorithm utilizes the BAG algorithm to acquire an initial solution, it is called TSDP-BAG. 

In Fig. \ref{HFL_latency_M8_N4}, we show the impacts of $d_{2}^{(2)}(t)$ on the HFL latency, when $M=8$ and $N=4$. When $d_{2}^{(2)}(t) \in \{10,20\}$, the TSDP-MaxSNR algorithm, the TSDP-BAG algorithm and the exhaustive search algorithm have the same HFL latency. On the other hand, when $d_{2}^{(2)}(t) \ge 30$, the TSDP-MaxSNR algorithm and the TSDP-BAG algorithm are inferior to the exhaustive search algorithm in terms of the HFL latency. It shows that the TSDP-assisted algorithm does not always produce an optimal matrix of user association when the MaxSNR algorithm or the BAG algorithm is used to obtain an initial solution. However, the TSDP-assisted algorithm is a polynomial-time algorithm but the exhaustive search algorithm is not. 

\begin{figure}
    \centering
    \includegraphics[width=7cm]{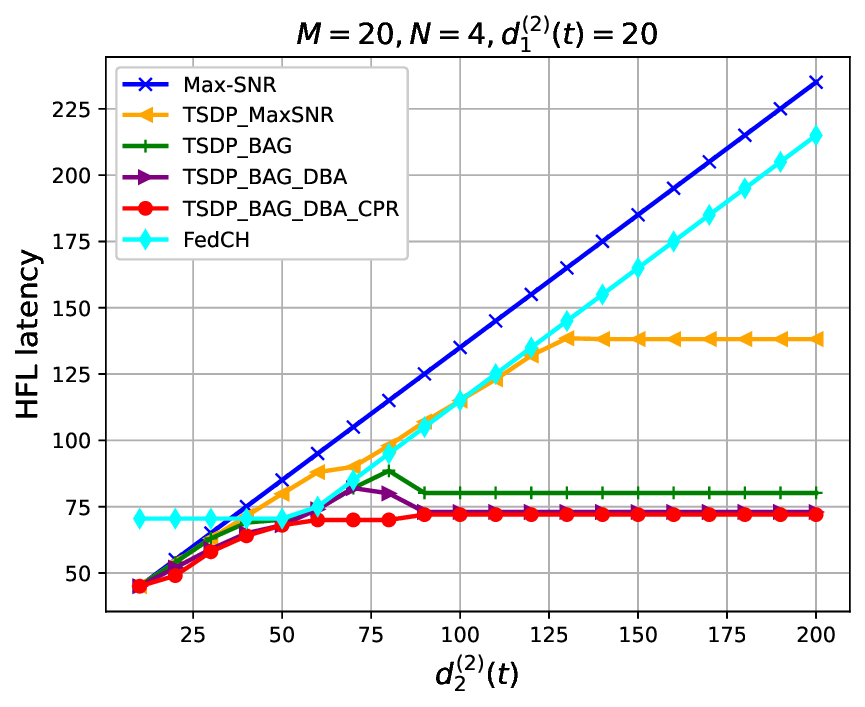}
    \caption{The HFL latency for six algorithms, $(M,N)=(20,4)$.}
    \label{HFL_latency_M20_N4_FedCH}
\end{figure}

\begin{figure}
    \centering
    \includegraphics[width=7cm]{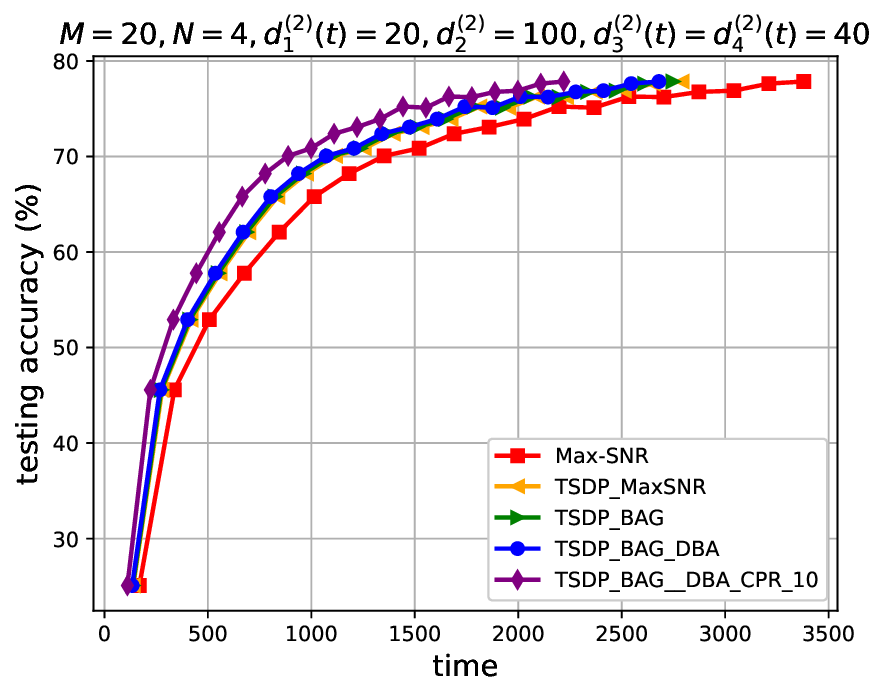}
    \caption{Testing accuracy of machine learning, when $M=20$.}
    \label{HFL_accuracy_M20_CPR10}
\end{figure}

\begin{figure}
    \centering
    \includegraphics[width=7cm]{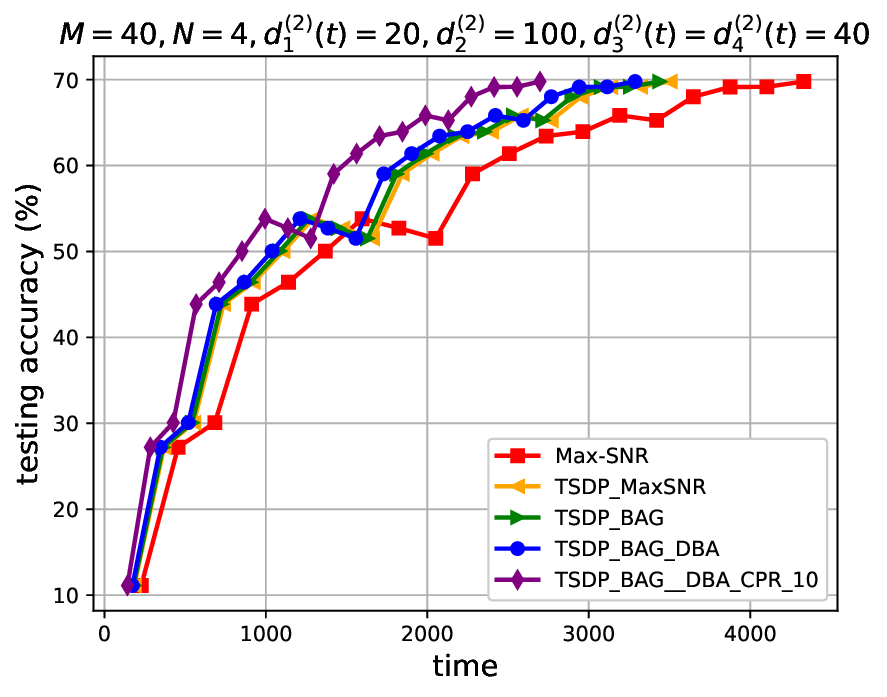}
    \caption{Testing accuracy of machine learning, when $M=40$.}
    \label{HFL_accuracy_M40_CPR10}
\end{figure}

We evaluate two variants of the TSDP-assisted algorithm with dynamic bandwidth allocation. The TSDP-BAG-DBA algorithm is composed of the first four phases of the TSDP-assisted algorithm and adopts the BAG algorithm in the first phase. The TSDP-BAG-DBA-CPR algorithm consists of the five phases of the TSDP-assisted algorithm, makes use of the BAG algorithm in the first phase and utilizes the CPR-M algorithm $10$ times in the fifth phase. We study the case in which different mobile users have distinct delays for uploading their local models to edge server $n$, $\forall n \in [N]$. Specifically, $M=40$, $N=4$, $K=10$, $\beta_{1,1}(t)=1$, $\beta_{2K+1,1}(t)=9$,  $\beta_{1,2}(t)=16$, $\beta_{2K+1,2}(t)=4$,  $\beta_{1,3}(t)=\beta_{2K+1,3}(t)=25=\beta_{1,4}(t)=\beta_{2K+1,4}(t)$, $\beta_{m,n}(t)=\beta_{1,n}(t)+0.1 \cdot (m-1)$, $\forall n \in [N], m \in [2K], t \in \mathbb{N}$ and $\beta_{2K+m,n}(t)=\beta_{2K+1,n}(t)+0.1 \cdot (m-1)$, $\forall n \in [N], m \in [2K], t \in \mathbb{N}$. 

In Fig. \ref{HFL_latency_M20_N4_FedCH}, we show the HFL latency for six algorithms when $(M,N)=(20,4)$. The network topology is the same as that for Fig. \ref{HFL_latency_M20_N2_FedCH} except that $N=4$, the third edge server is at $(-20,10)$ and the fourth edge server is at $(20,10)$. Among the six algorithms, the TSDP-BAG-DBA-CPR algorithm is the best in terms of the HFL latency when $d_{2}^{(2)}(t) \in [10, 200]$. When $d_{2}^{(2)}(t) \in [40, 200]$, the FedCH algorithm is superior to the Max-SNR algorithm. On the other hand, when $d_{2}^{(2)}(t) \in [10, 30]$, the FedCH algorithm is inferior to the Max-SNR algorithm. When $d_{2}^{(2)}(t) \in [50, 80]$, the FedCH algorithm outperforms the TSDP-MaxSNR algorithm. When $d_{2}^{(2)}(t) \in [130, 200]$, the FedCH is worse than the TSDP-MaxSNR algorithm. When $d_{2}^{(2)}(t) \in [90, 200]$, the HFL latency of the TSDP-BAG-DBA-CPR algorithm is slighter smaller than that of the TSDP-BAG-DBA algorithm.

We evaluate the testing accuracy of hierarchical federated learning when $M$ mobile users have independent and identically distributed (IID) training data items. In addition, $M$ mobile users update their local models in each round. In Fig. \ref{HFL_accuracy_M20_CPR10}, we show the testing accuracy of machine learning for five studied algorithms when $M=20$ and $N=4$. Among the studied algorithms, the TSDP-BAG-DBA-CPR algorithm has the maximum convergence speed and the TSDP-BAG-DBA algorithm has the second largest convergence speed. In comparison with the Max-SNR algorithm, the proposed TSDP-assisted algorithm could reduce the length of an HFL round. In addition, dynamic bandwidth allocation reduces the length of an HFL round. After $15$ rounds, the TSDP-BAG-DBA-CPR algorithm converges and the corresponding testing accuracy is larger than $76\%$. In Fig. \ref{HFL_accuracy_M40_CPR10}, we show the testing accuracy of machine learning for five studied algorithms when $M=40$ and $N=4$. In this case, the maximum testing accuracy is $69.78\%$. Since the number of training data items is fixed, as the value of $M$ increases, each mobile user has fewer training data items. Thus, as $M$ increases from $20$ to $40$, the testing accuracy slightly decreases. 


\section{Conclusion}

We have proposed novel algorithms for user association and wireless bandwidth allocation for a hierarchical federated learning system in which mobile devices have unequal computational capabilities and edge servers have different model uploading delays to the cloud server. To minimize the length of a global round by optimal user association, we have formulated a combinatorial optimization problem. We have designed the twin sorting dynamic programming algorithm that obtains an optimal user association matrix in polynomial time when there are two edge servers and equal bandwidth allocation is adopted. In addition, we have proposed the TSDP-assisted algorithm that could exploit the TSDP algorithm for efficiently obtaining an appropriate matrix of user association when there are three or more edge servers. Given a user association matrix, to attain optimal wireless bandwidth allocation, we have formulated and solved a convex optimization problem. The TSDP-assisted algorithm also utilizes critical path reduction to reduce the HFL latency. We have used simulation results to reveal the advantages of the proposed approach over a number of alternative schemes. Future work includes jointly optimizing user-edge association and wireless resource allocation for hierarchical federated learning with device-to-device communications.


\begin{thebibliography}{1}   

\bibitem{McMahan2017} H. B. McMahan, E. Moore, D. Ramage, S. Hampson, and B. A. Y. Arcas, ``Communication-Efficient Learning of Deep Networks from Decentralized Data," in \textit{Proc. 2017 International Conference on Artificial Intelligence and Statistics (AISTATS)}, Ft. Lauderdale, FL, USA, Apr. 20-22, 2017.

\bibitem{Abad2020ICASSP} M. S. H. Abad, E. Ozfatura, D. GUndUz and O. Ercetin, ``Hierarchical Federated Learning ACROSS Heterogeneous Cellular Networks," in \textit{Proc. 2020 IEEE International Conference on Acoustics, Speech and Signal Processing (ICASSP)}, pp. 8866-8870, Barcelona, Spain, May 4-8, 2020.

\bibitem{Liu2020ICC} L. Liu, J. Zhang, S. H. Song and K. B. Letaief, ``Client-Edge-Cloud Hierarchical Federated Learning," in \textit{Proc. 2020 IEEE International Conference on Communications (ICC)}, pp. 1-6, Dublin, Ireland, June 7-11, 2020.

\bibitem{Luo2020TWC} S. Luo, X. Chen, Q. Wu, Z. Zhou and S. Yu, ``HFEL: Joint Edge Association and Resource Allocation for Cost-Efficient Hierarchical Federated Edge Learning," \textit{IEEE Trans. Wireless Commun.}, vol. 19, no. 10, pp. 6535-6548, Oct. 2020.

\bibitem{Liu2022TWC} S. Liu, G. Yu, X. Chen and M. Bennis, ``Joint User Association and Resource Allocation for Wireless Hierarchical Federated Learning With IID and Non-IID Data," \textit{IEEE Trans. Wireless Commun.}, vol. 21, no. 10, pp. 7852-7866, Oct. 2022.

\bibitem{Liu2022SEC} C. Liu, T. J. Chua and J. Zhao, ``Time Minimization in Hierarchical Federated Learning," in \textit{Proc. 2022 IEEE/ACM 7th Symposium on Edge Computing (SEC)}, pp. 96-106, Seattle, WA, USA, Dec. 2022.

\bibitem{Gau2024VTC} R.-H. Gau, D.-C. Liang, T.-Y. Wang and C.-H. Liu, ``Edge-to-cloud Latency Aware User Association in Wireless Hierarchical Federated Learning," in \textit{Proc. 2024 IEEE Vehicular Technology Conference (VTC2024-Spring)}, Singapore, June 24-27, 2024.

\bibitem{Wu2023TPDS} Q. Wu et al., ``HiFlash: Communication-Efficient Hierarchical Federated Learning With Adaptive Staleness Control and Heterogeneity-Aware Client-Edge Association," \textit{IEEE Trans. Parallel Distrib. Syst.}, vol. 34, no. 5, pp. 1560-1579, May 2023.

\bibitem{Wang2023TMC} Z. Wang, H. Xu, J. Liu, Y. Xu, H. Huang and Y. Zhao, ``Accelerating Federated Learning With Cluster Construction and Hierarchical Aggregation," \textit{IEEE Trans. Mobile Comput.}, vol. 22, no. 7, pp. 3805-3822, July 2023.

\bibitem{Deng2024ToN} Y. Deng \textit{et al.}, ``A Communication-Efficient Hierarchical Federated Learning Framework via Shaping Data Distribution at Edge," \textit{IEEE/ACM Trans. Netw.}, vol. 32, no. 3, pp. 2600-2615, June 2024.

\bibitem{Dinh2021ToN} C. T. Dinh \textit{et al.}, ``Federated Learning Over Wireless Networks: Convergence Analysis and Resource Allocation," \textit{IEEE/ACM Trans. Netw.}, vol. 29, no. 1, pp. 398-409, Feb. 2021.

\bibitem{Chen2021TWC} M. Chen, H. V. Poor, W. Saad and S. Cui, ``Convergence Time Optimization for Federated Learning Over Wireless Networks," \textit{IEEE Trans. Wireless Commun.}, vol. 20, no. 4, pp. 2457-2471, April 2021.

\bibitem{Wen2022TWC} W. Wen, Z. Chen, H. H. Yang, W. Xia and T. Q. S. Quek, ``Joint Scheduling and Resource Allocation for Hierarchical Federated Edge Learning," \textit{IEEE Trans. Wireless Commun.}, vol. 21, no. 8, pp. 5857-5872, Aug. 2022.
   
\bibitem{Liu2023TWC} L. Liu, J. Zhang, S. Song and K. B. Letaief, ``Hierarchical Federated Learning With Quantization: Convergence Analysis and System Design," \textit{IEEE Trans. Wireless Commun.}, vol. 22, no. 1, pp. 2-18, Jan. 2023.

\bibitem{Chen2023TVT} Q. Chen, Z. You, D. Wen and Z. Zhang, ``Enhanced Hybrid Hierarchical Federated Edge Learning Over Heterogeneous Networks," \textit{IEEE Trans. Veh. Technol.}, vol. 72, no. 11, pp. 14601-14614, Nov. 2023.

\bibitem{Feng2022TWC} C. Feng, H. H. Yang, D. Hu, Z. Zhao, T. Q. S. Quek and G. Min, ``Mobility-Aware Cluster Federated Learning in Hierarchical Wireless Networks," \textit{IEEE Trans. Wireless Commun.}, vol. 21, no. 10, pp. 8441-8458, Oct. 2022.


\bibitem{Fooladivanda2013TWC} D. Fooladivanda and C. Rosenberg, ``Joint Resource Allocation and User Association for Heterogeneous Wireless Cellular Networks," \textit{IEEE Trans. Wireless Commun.}, vol. 12, no. 1, pp. 248-257, January 2013.

\bibitem{Ye2013TWC} Q. Ye, B. Rong, Y. Chen, M. Al-Shalash, C. Caramanis and J. G. Andrews, ``User Association for Load Balancing in Heterogeneous Cellular Networks," \textit{IEEE Trans. Wireless Commun.}, vol. 12, no. 6, pp. 2706-2716, June 2013.

\bibitem{Lin2015JSAC} Y. Lin, W. Bao, W. Yu and B. Liang, ``Optimizing User Association and Spectrum Allocation in HetNets: A Utility Perspective," \textit{IEEE J. Sel. Areas Commun.}, vol. 33, no. 6, pp. 1025-1039, June 2015.

\bibitem{Oo2017TMC} T. Z. Oo, N. H. Tran, W. Saad, D. Niyato, Z. Han and C. S. Hong, ``Offloading in HetNet: A Coordination of Interference Mitigation, User Association, and Resource Allocation," \textit{IEEE Trans. Mobile Comput.}, vol. 16, no. 8, pp. 2276-2291, Aug. 2017.

\bibitem{Zhao2019TWC} N. Zhao, Y.-C. Liang, D. Niyato, Y. Pei, M. Wu and Y. Jiang, ``Deep Reinforcement Learning for User Association and Resource Allocation in Heterogeneous Cellular Networks," \textit{IEEE Trans. Wireless Commun.}, vol. 18, no. 11, pp. 5141-5152, Nov. 2019.

\bibitem{Luo2015TWC} S. Luo, R. Zhang and T. J. Lim, ``Downlink and Uplink Energy Minimization Through User Association and Beamforming in C-RAN," \textit{IEEE Trans. Wireless Commun.}, vol. 14, no. 1, pp. 494-508, Jan. 2015.

\bibitem{Chien2016TWC} T. Van Chien, E. Bjornson and E. G. Larsson, ``Joint Power Allocation and User Association Optimization for Massive MIMO Systems," \textit{IEEE Trans. Wireless Commun.}, vol. 15, no. 9, pp. 6384-6399, Sept. 2016.

\bibitem{Sekander2017TMC} S. Sekander, H. Tabassum and E. Hossain, ``Decoupled Uplink-Downlink User Association in Multi-Tier Full-Duplex Cellular Networks: A Two-Sided Matching Game," \textit{IEEE Trans. Mobile Comput.}, vol. 16, no. 10, pp. 2778-2791, Oct. 2017.

\bibitem{Khawam2022TMC} K. Khawam, S. Lahoud, M. E. Helou, S. Martin and F. Gang, ``Coordinated Framework for Spectrum Allocation and User Association in 5G HetNets With mmWave," \textit{IEEE Trans. Mobile Comput.}, vol. 21, no. 4, pp. 1226-1243, April 2022.

\bibitem{Zarifneshat2023TMC} M. Zarifneshat, P. Roy and L. Xiao, ``Multi-Objective Approach for User Association to Improve Load Balancing and Blockage in Millimeter Wave Cellular Networks," \textit{IEEE Trans. Mobile Comput.}, vol. 22, no. 5, pp. 2818-2836, May 2023.

\bibitem{Huang2022TMC} X. Huang, S. Zhao, X. Gao, Z. Shao, H. Qian and Y. Yang, ``Online User-AP Association With Predictive Scheduling in Wireless Caching Networks," \textit{IEEE Trans. Mobile Comput.}, vol. 21, no. 6, pp. 2116-2129, June 2022.

\bibitem{Li2022TMC} Y. Li, H. Ma, L. Wang, S. Mao and G. Wang, ``Optimized Content Caching and User Association for Edge Computing in Densely Deployed Heterogeneous Networks," \textit{IEEE Trans. Mobile Comput.}, vol. 21, no. 6, pp. 2130-2142, June 2022.

\bibitem{Chen2023TMC} W.-Y. Chen, P.-Y. Chou, C.-Y. Wang, R.-H. Hwang and W.-T. Chen, ``Dual Pricing Optimization for Live Video Streaming in Mobile Edge Computing With Joint User Association and Resource Management," \textit{IEEE Trans. Mobile Comput.}, vol. 22, no. 2, pp. 858-873, Feb. 2023.

\bibitem{Nouri2023TMC} N. Nouri, F. Fazel, J. Abouei and K. N. Plataniotis, ``Multi-UAV Placement and User Association in Uplink MIMO Ultra-Dense Wireless Networks," \textit{IEEE Trans. Mobile Comput.}, vol. 22, no. 3, pp. 1615-1632, Mar. 2023.

\bibitem{Dai2023TMC} C. Dai, K. Zhu and E. Hossain, ``Multi-Agent Deep Reinforcement Learning for Joint Decoupled User Association and Trajectory Design in Full-Duplex Multi-UAV Networks," \textit{IEEE Trans. Mobile Comput.}, vol. 22, no. 10, pp. 6056-6070, Oct. 2023.

\bibitem{Liu2020TVT} Y.-J. Liu, G. Feng, Y. Sun, S. Qin and Y.-C. Liang, ``Device Association for RAN Slicing Based on Hybrid Federated Deep Reinforcement Learning," \textit{IEEE Trans. Veh. Technol.}, vol. 69, no. 12, pp. 15731-15745, Dec. 2020.

\bibitem{Lim2021JSAC} W. Y. B. Lim, J. S. Ng, Z. Xiong, D. Niyato, C. Miao and D. I. Kim, ``Dynamic Edge Association and Resource Allocation in Self-Organizing Hierarchical Federated Learning Networks," \textit{IEEE J. Sel. Areas Commun.}, vol. 39, no. 12, pp. 3640-3653, Dec. 2021.
 
\bibitem{Li2022TVT} Y. Li, X. Qin, H. Chen, K. Han and P. Zhang, ``Energy-Aware Edge Association for Cluster-Based Personalized Federated Learning," \textit{IEEE Trans. Veh. Technol.}, vol. 71, no. 6, pp. 6756-6761, June 2022.

\bibitem{Lin2023TVT} Y. Lin, J. Bao, Y. Zhang, J. Li, F. Shu and L. Hanzo, ``Privacy-Preserving Joint Edge Association and Power Optimization for the Internet of Vehicles via Federated Multi-Agent Reinforcement Learning," \textit{IEEE Trans. Veh. Technol.}, vol. 72, no. 6, pp. 8256-8261, June 2023.

\bibitem{Ong2022ICC} S.-C. Ong and R.-H. Gau, ``Local Loss-Assisted Dynamic Client Selection for Image Classification-Oriented Federated Learning," in \textit{Proc. 2022 IEEE International Conference on Communications (ICC)}, pp. 4769-4774, Seoul, Republic of Korea, May 16-20, 2022.  

\bibitem{Hsu2023VTC} Y.-S. Hsu and R.-H. Gau, ``MMSE Threshold-based Power Control for Wireless Federated Learning," in \textit{Proc. 2023 IEEE Vehicular Technology Conference (VTC2023-Spring)}, pp. 1-6, Florence, Italy, June 20-23, 2023.

\bibitem{Hosseinalipour2022ToN} S. Hosseinalipour \textit{et al.}, ``Multi-Stage Hybrid Federated Learning Over Large-Scale D2D-Enabled Fog Networks," \textit{IEEE/ACM Trans. Netw.}, vol. 30, no. 4, pp. 1569-1584, Aug. 2022.

\bibitem{Ganguly2023ToN} B. Ganguly \textit{et al.}, ``Multi-Edge Server-Assisted Dynamic Federated Learning With an Optimized Floating Aggregation Point," \textit{IEEE/ACM Trans. Netw.}, vol. 31, no. 6, pp. 2682-2697, Dec. 2023.

\bibitem{Liu2024ToN} J. Liu, J. Liu, H. Xu, Y. Liao, Z. Wang and Q. Ma, ``YOGA: Adaptive Layer-Wise Model Aggregation for Decentralized Federated Learning," \textit{IEEE/ACM Trans. Netw.}, vol. 32, no. 2, pp. 1768-1780, April 2024.

\bibitem{Lim2020Survey} W. Y. B. Lim \textit{et al.}, ``Federated Learning in Mobile Edge Networks: A Comprehensive Survey," \textit{IEEE Communications Surveys and Tutorials}, vol. 22, no. 3, pp. 2031-2063, third quarter 2020.
 
\bibitem{Kar2023Survey} B. Kar, W. Yahya, Y. -D. Lin and A. Ali, ``Offloading Using Traditional Optimization and Machine Learning in Federated Cloud–Edge–Fog Systems: A Survey," \textit{IEEE Communications Surveys and Tutorials}, vol. 25, no. 2, pp. 1199-1226, Second quarter 2023.

\bibitem{CVXPY} S. Diamond and S. Boyd, ``CVXPY: A Python-embedded modeling language for convex optimization," \textit{Journal of Machine Learning Research}, vol. 17, no. 83, pp. 1-5, 2016.

\bibitem{Algorithms} T. H. Cormen, C. E. Leiserson, R. L. Rivest and C. Stein, \textit{Introduction to Algorithms}, 4th Edition. Cambridge, MA, USA: The MIT Press, 2022.

\bibitem{PyTorch} A. Paszke \textit{et al.}, ``PyTorch: An Imperative Style, High-Performance Deep Learning Library," in \textit{Proc. 2019 Conference on Neural Information Processing Systems (NeurIPS)}, Vancouver, Canada, Dec. 10-12, 2019.

\bibitem{CIFAR-10} Alex Krizhevsky, ``Learning Multiple Layers of Features from Tiny Images," 2009, https://www.cs.toronto.edu/$\sim$kriz/learning-features-2009-TR.pdf.

\end{thebibliography}
\end{document}